\documentclass{article}

\usepackage[letterpaper]{geometry}
\usepackage[parfill]{parskip}
\PassOptionsToPackage{numbers, compress}{natbib}
\usepackage[numbers, compress]{natbib}
\usepackage{xr-hyper,refcount}

\usepackage{graphicx}
\usepackage[utf8]{inputenc} 
\usepackage[T1]{fontenc}    
\usepackage{url}            
\usepackage{booktabs}       
\usepackage{amsfonts}       
\usepackage{nicefrac}       
\usepackage{microtype}      
\usepackage[parfill]{parskip}

\usepackage{amsmath,amsthm,amssymb,bbm}
\usepackage{mathtools}
\usepackage{cases}
\usepackage{comment}
\usepackage{subcaption}

\usepackage{color}
\usepackage{appendix}
\usepackage{xspace}
\usepackage{enumitem}
\usepackage[english]{babel}
\usepackage{authblk}

\usepackage[colorlinks,citecolor=blue,urlcolor=blue,linkcolor=blue,linktocpage=true]{hyperref}
\usepackage{cleveref}
\crefformat{equation}{(#2#1#3)}
\crefrangeformat{equation}{(#3#1#4) to~(#5#2#6)}
\crefname{equation}{}{}
\Crefname{equation}{}{}

\crefname{definition}{\textbf{definition}}{definitions}
\Crefname{definition}{Definition}{Definitions}
\crefname{assumption}{\textbf{assumption}}{assumptions}
\Crefname{assumption}{Assumption}{Assumptions}
\definecolor{maroon}{RGB}{192,80,77}

\newtheorem{theorem}{Theorem}
\newtheorem{lemma}[theorem]{Lemma}
\newtheorem{fact}[theorem]{Fact}

\newtheorem{corollary}[theorem]{Corollary}

\newtheorem{remark}[theorem]{Remark}

\DeclarePairedDelimiter\ceil{\lceil}{\rceil}
\DeclarePairedDelimiter\floor{\lfloor}{\rfloor}

\DeclareMathOperator\E{E}

\usepackage[utf8]{inputenc} 
\usepackage[T1]{fontenc}    
\usepackage{lmodern}
\usepackage{hyperref}       
\usepackage{url}            
\usepackage{booktabs}       
\usepackage{amsfonts}       
\usepackage{nicefrac}       
\usepackage{microtype}      

\usepackage{natbib}
\usepackage{amsmath}
\usepackage{bbm}

\usepackage[normalem]{ulem}
\usepackage[ruled]{algorithm2e}

\usepackage{mathtools}
\usepackage{thm-restate}
\usepackage{amssymb,amsmath,amsthm}

\newcommand{\BlackBox}{\rule{1.5ex}{1.5ex}}  
\newenvironment{proof-sketch}{\par\noindent{\bf Proof Sketch\ }}{\hfill\BlackBox\\[2mm]}
\long\def\acks#1{\vskip 0.3in\noindent{\large\bf Acknowledgments}\vskip 0.2in
\noindent #1}

\title{Thompson Sampling for Adversarial Bit Prediction}

\begin{document}

\author{
  Yuval Lewi\thanks{Tel Aviv University. Email: \href{mailto:yuval.lewi@gmail.com}{\nolinkurl{yuval.lewi@gmail.com}}.}\hspace*{1cm}
  Haim Kaplan\thanks{Tel Aviv University and Google Research. Email: \href{mailto:haimk@tau.ac.il}{\nolinkurl{haimk@tau.ac.il}}.}\hspace*{1cm}
  Yishay Mansour\thanks{Tel Aviv University and Google Research. Email: \href{mailto:mansour.yishay@gmail.com}{\nolinkurl{mansour.yishay@gmail.com}}.}
}

\maketitle

\begin{abstract}
We study the Thompson sampling algorithm in an adversarial setting, specifically, for adversarial bit prediction.
We characterize the bit sequences with the smallest and largest expected regret. Among sequences of length $T$
with $k < \frac{T}{2}$ zeros, the  sequences of
largest regret consist of alternating  zeros and ones followed by the remaining ones, and
the sequence  of smallest regret consists of  ones followed by zeros.
We also bound the regret of those sequences,
the worst case sequences have regret  $O(\sqrt{T})$
and the best case sequence have regret  $O(1)$.

We extend our results to a model where false positive and false negative errors have different weights.
We characterize the sequences with largest  expected regret in this generalized setting, and derive their regret bounds. We also show that there are sequences with $O(1)$ regret.
\end{abstract}

\section{Introduction}

Online learning and multi-arm bandits (MAB) are one of the most basic models for
uncertainty, which are widely studied in machine learning.
The main performance criteria used in this model is regret, which is
the difference between the expected loss of the online algorithm, and the loss of the best algorithm from a benchmark class.
(See, \cite{cesa2006prediction,BubeckC12,book-TLCS,book-AS}).
Bit prediction is one of the first problems  for which online learning regret was analyzed \citep{Cover1966}, 
and has been extensively studied throughout the years (see, \cite{book-ARKS}).

Thompson sampling (\cite{thompson1933})  is 
one of the earliest algorithms for  MAB. It was originally motivated by a
Bayesian setting, where the rewards are stochastic, and the reward
of each action has a prior distribution. The algorithm maintains a
posterior distribution for the reward of each action, and in each
step, samples the posterior distribution of the mean reward of each
action, and uses the action with the highest sampled value. In
recent years, there has been a renewed interest in the Thompson
sampling algorithm and its applications (see,
\cite{russo2018tutorial}), mainly due to its simplicity and good performance in
practice.

Since Thompson sampling was designed for a Bayesian setting, it is natural to analyze its   Bayesian regret (i.e.,\ average the regret with respect to the prior).
In many settings, we get an elegant analysis and asymptotically optimal regret bounds.
(See, \cite{book-TLCS,book-AS,RussoR16}).

While Thompson sampling was designed for a Bayesian setting, it was also recently analyzed  in  worst-case
stochastic setting.
More specifically, assume that the reward of each action is a Bernoulli random variable with unknown success probability.
Unlike the Bayesian setting, there is no true prior over these parameters (success probabilities),
and we want to bound the regret for the worst choice of the parameters.
In this setting we start
the Thompson sampling algorithm  with a {\em fictitious}  prior, say, a uniform distribution (of the success probability) for each action, and
we update the posterior as though we were in the Bayesian setting.
The works of \cite{agrawal2017near,agrawal2013further} show that Thompson sampling guarantees  almost optimal regret bounds in the adversarial stochastic setting. Improved regret bounds which are parameter dependent are given in \cite{kaufmann2012thompson}.

The papers mentioned above show the great success of Thompson sampling in stochastic settings, thus it is natural to investigate its performance in adversarial online model.
In this model TS starts with a fictitious prior and an adversary selects the arbitrary input sequence. 
The completely adversatial model can be viewed as bounding the regret of the worst-case sequence possible, rather then the expected regret over some distribution in the stochastic settings.
Specifically in this paper, our goal is to show that Thompson Sampling is successful for the adversarial bit sequence settings.

Our work considers the performance of Thompson sampling in an  adversarial setting.
Specifically, we consider the case of adversarial bit prediction,
where the learner observes an arbitrary binary sequence, and at each time step predicts the next bit. The loss of the learner is the
number of errors it makes, and the regret is the difference between
the number of errors the online learner algorithm makes and  best static bit
prediction, i.e., the minimum between the number of ones and zeros
in the sequence. 
We characterize the bit sequences on which Thompson sampling algorithm has the
largest and smallest regret. We  bound the regret of these sequences, and show that the worst case regret is $\Theta(\sqrt{T})$, for a sequence of length $T$, and best case regret of $\Theta(1)$.

More specifically, we initialize our Thompson sampling algorithm with a uniform (i.e., $\beta(1,1)$) prior distribution, and
 maintain a posterior beta distribution (whose parameters correspond to the number of ones and zeros seen so far).
To predict the next bit, we draw a value from the beta posterior and predict one if the value is larger than $\frac{1}{2}$.
Once we observe the bit we update our posterior.

For sequences of length $T$ with $k\leq \frac{T}{2}$ zeros, we show
that the sequences with the largest regret are of the form
$\{01,10\}^k 1^{T-2k}$, and the sequence with the smallest regret is
$1^{T-k}0^k$ (for $k=\frac{T}{2}$  both sequences $1^{T/2}0^{T/2}$
and $0^{T/2}1^{T/2}$ have the same smallest regret). For example, if $k=2$ and $T=7$, the sequences with the largest regret are $0101111, 0110111,1001111$ and $1010111$, and the sequence with the smallest regret is $1111100$. For
$k>\frac{T}{2}$, we have the same characterization with $1$ and $0$
interchanged.
We also bound the regret of these sequences and show that the
expected regret on the worst case sequences is  $\Theta(\sqrt{T})$
and that the expected regret on the best case sequences is
$\Theta(1)$.

We extend the model to have different losses for false positive and
false negative errors. Specifically, we have a trade-off parameter
$q\in[0,1]$ and we define the cost of a false positive to be $q$ and
the cost of a false negative to be $1-q$. We call this extended
model the \textit{generalized bit-prediction} model. Note that for
$q=\frac{1}{2}$ this loss is simply the number of errors multiplied
by $\frac{1}{2}$, so this is a strict generalization our previous
loss. Thompson sampling adapts naturally to the
parameter $q$, by simply predicting one when the sampled value is
larger than $q$ (rather than larger than $\frac{1}{2}$).
We characterize for each $q \in [0,1]$ the bit sequences with the
largest regret for this model and bound their regret. For example,
for sequences of length $T=100$ with $20$ zeros and $q=\frac{1}{3}$,
the worst case sequences are of the form $\{010,001\}^{10} 1^{70}$.
In general, we show a family of bit-sequences with the highest
regret for every trade-off parameter $q \in [0,1]$, number of zeros
and number of ones. From that we conclude that the regret of
Thompson sampling in the adversarial bit-prediction model is bounded by
$O(\sqrt{q(1-q)T})$. We also show that there are sequences with regret equals or less then $1$ without characterizing the best sequences.

Our work shows the great versatility of Thompson sampling. Namely,
the same algorithm, with a prior of $\beta(1,1)$, can be analysed in
Bayesian setting, when it is given the true prior, in an adversarial
stochastic setting, when it is given a fictitious prior, and in the
adversarial bit prediction problem, which we analyse in this work.
Thompson sampling is not the only algorithm that achieves good
performance both for adversarial and stochastic rewards (See,
\cite{BubeckS12,SeldinS14,Mourtada18}), but it achieves this in a
simple natural way, and as a side-product of a general Bayesian
methodology, without trying to identify the nature of the
environment.

\subsection{Other related work}

Adversarial bit prediction has a long history, starting with
\cite{Cover1966}, and followed up by many additional works (see,
\cite{cesa2006prediction}). The exact min-max optimal strategy can
be derived, when we view the problem as a zero-sum game (see,
\cite{book-ARKS}). The min-max optimal regret bound for the case of
two actions was derived by \cite{Cover1966} and for three actions by
\cite{GravinPS16}.
Prediction of the next character in non-binary
sequences has also received considerable attention, with respect to
various benchmarks \cite{FederMG92,cesa1998individualsequences}.
For the stochastic case, prediction of the next character in non-binary
sequences was studied using Bayesian methods by \cite{hutter2003optimality}.
Prediction of
binary sequences with the log-loss in online adversarial environment has been studied by many due to its relation to data compression and information-theory (see for example, \cite{freund1996predicting}, \cite{merhav1998universal} and \cite{xie2000asymptotic}).

Adversarial online learning and multi-arm bandits  have received significant attention in machine learning in the last two decades.
(See the following books and surveys,  \cite{cesa2006prediction,BubeckC12,book-TLCS,book-AS}).
A lower bound for the adversarial MAB problem was presented by \cite{seldin2016lower}.
Notable results in adversarial online learning are the algorithm EXP3 (see, \cite{auer2002nonstochastic}) for adversarial bandits, the algorithm UCB1 (see, \cite{auer2002finite}) for stochastic bandits, and the regret analysis of the min-max algorithm (see, \cite{audibert2009minimax}).

Thompson sampling has been studied in different environments over the years. In \cite{Gopalan13} it was observed that Thompson sampling with a
Gaussian prior is equivalent to "Follow the Perturbed Leader" (FPL) of
\cite{KalaiV05}, and that fact was used to deduced the worst case regret of Thompson sampling with Gaussian distributions. A prior-dependent analysis was introduced by
\cite{RussoR16} using an information-theoretic tools, and the idea was expanded for first and second-order regret  bounds by \cite{bubeck2019first}.

Thompson sampling also showed good experimental results (see, \cite{scott2010modern,chapelle2011empirical}). Because of that, the algorithm is used in practice, with recommendation systems as an example (see, \cite{kawale2015efficient}). In Reinforcement Learning, a version of Thompson sampling called "Posterior Sampling for Reinforcement Learning" (PSRL) is used (see, \cite{osband2013more, osband2017posterior}). Bounds for the algorithm were proved in \cite{agrawal2017optimistic}.

\section{Model}\label{sec:model}


A \textit{bit prediction} game proceeds as follows. At time $t\in
\left[ T \right]=\lbrace 1,...,T \rbrace$ the learner outputs a bit
$\hat{\gamma}_t \in \lbrace 0,1 \rbrace$. Then, the learner observes
a bit $\gamma_t \in \lbrace 0,1 \rbrace$ and suffers a loss of $\ell
\left( \hat{\gamma}_t, \gamma_t\right) = \mathbb{I} \lbrace
\hat{\gamma}_t \neq \gamma_t \rbrace$.


We compare the loss of the online algorithm to a benchmark, which is the loss of
the best static bit prediction.
Given a bit sequence $\Gamma= \left(\gamma_1,..., \gamma_T \right)$,
let the number of ones up to $t$ be
$O_t\left(\Gamma\right)=\lvert
\left\lbrace i\in \left[ t \right]: \gamma_i=1 \right\rbrace \rvert
= \sum_{i=1}^t \gamma_i$ and the number of zeros be
$Z_t\left(\Gamma\right)= \lvert \left\lbrace i\in \left[ t \right]:
\gamma_i=0 \right\rbrace \rvert $ $= \sum_{i=1}^t \left(1-\gamma_i
\right)$.
The loss of the best static bit prediction is
\[
static \left( \Gamma \right) = \min \left\lbrace \sum_{t=1}^T \ell \left(
1, \gamma_t \right), \sum_{t=1}^T \ell \left( 0, \gamma_t \right)
\right\rbrace = \min \left\lbrace Z_T \left( \Gamma \right), O_T \left( \Gamma
\right) \right\rbrace.
\]

The goal of the learner is to minimize the regret, which is the difference
between the online cumulative loss and the loss of the best static
bit prediction. Specifically, for an algorithm $A$,
\[
Regret_A ( \Gamma ) = \sum_{t=1}^{T} E_{\hat{\gamma}_t \sim A}  [ \ell(\hat{\gamma}_t,
\gamma_t) \mid \Gamma ] - static ( \Gamma ),
\]
where $\Gamma \in \{ 0,1 \}^T$ is a fixed bit sequence, and the
expectation is taken over the predictions of algorithm $A$. We
extend the standard bit prediction game and define a
\textit{generalized bit prediction} game, where the false positive
(FP) and false negative (FN) errors have different
weights.\footnote{A false positive error is when the learner
predicts $\hat{\gamma}_t=1$ and $\gamma_t=0$, and false negative
error is when $\hat{\gamma}_t=0$ and $\gamma_t=1$.}
Given a \textit{trade-off parameter} $q \in [0,1]$, we define a loss
$\ell^q$, as follows,
\begin{align*}
\ell^q \left( \hat{\gamma}_t, \gamma_t \right) &=  q \mathbb{I}
\lbrace \hat{\gamma}_t =1,\gamma_t=0 \rbrace+ (1-q)\mathbb{I}
\lbrace \hat{\gamma}_t =0 , \gamma_t=1 \rbrace.
\end{align*}
Namely, the false positive errors are weighted by $q$ while the
false negative errors are weighted by $1-q$.
Note that for $q=\frac{1}{2}$, for any $( \hat{\gamma}_t, \gamma_t)$
we have that $\ell^{1/2}( \hat{\gamma}_t, \gamma_t )=\frac{1}{2}\ell(
\hat{\gamma}_t, \gamma_t )$, so for $q=\frac{1}{2}$ the extended loss is essentially the 0-1 loss.


Similarly, the benchmark for the generalized bit prediction is the
best static bit prediction, namely,
\[
static^q ( \Gamma ) = \min \left\lbrace \sum_{t=1}^T \ell^q ( 1, \gamma_t
) , \sum_{t=1}^T \ell^q ( 0, \gamma_t) \right\rbrace = \min \lbrace q Z_T(
\Gamma), (1-q) O_T ( \Gamma ) \rbrace,
\]
and the regret of algorithm $A$ on a given bit sequence $\Gamma \in \{ 0,1 \}^T$ is
\[
Regret_A^q ( \Gamma ) = \sum\limits_{t=1}^{T} E_{\hat{\gamma}_t \sim A} \left[
\ell^q(\hat{\gamma}_t, \gamma_t) \mid \Gamma \right] - static^q
\left( \Gamma \right).
\]

\subsection{Distributions}
\label{sec:mathematical-background}

We use extensively the Beta distribution, denoted by $\beta(a,b)$,
where $a,b>0$, and the Binomial distribution, denoted by $Bin(n,p)$
where $n$ is the number of trials and $p\in [0,1]$ is the success
probability. We  denote by $Ber(p)$ a Bernoulli random variable with
success probability $p\in[0,1]$. For a distribution $D$, the
Cumulative Distribution Function (CDF) is denoted by $F_D$.

The following identity is a well known fact related to the the Beta distribution (see, \cite{NIST:DLMF},
Eq.~8.17.4)

\begin{fact}\label{fact:beta-symmetry}
For $a,b \in \mathbb{N^+}$ and $p \in [0,1]$ we have
$F_{\beta(a,b)}(p)=1-F_{\beta(b,a)}(1-p)$.
\end{fact}

The $\beta(a,b)$ distribution is widely used in Bayesian setting to
define the uncertainty over the parameter $p$ of a Bernoulli random
variable $Ber(p)$. The distribution  $\beta(1,1)$, which is
the uniform distribution over $[0,1]$, is used as the prior distribution of $p$. Given
$a+b$ observations of the random variable $Ber(p)$, where $a$ is the
number of realizations  which are $1$ and $b$ is the number of realizations which are $0$,
then the posterior distribution of $p$ is $\beta(a+1,b+1)$ (assuming the
prior distribution is $\beta(1,1)$).

The following is a well known property of the CDF of the Beta
distribution.
\begin{fact}
\label{fact:RBF_properties} \cite[Eq.~8.17.20-21]{NIST:DLMF} For every $x
\in \left[ 0,1 \right]$ and $a,b \in \mathbb{R}$ s.t. $a,b>0$, the
following holds
\[
F_{\beta(a+1,b)}(x)=F_{\beta(a,b)}-\frac{x^{a}(1-x)^{b}}{a\mathrm{B}
\left(a,b \right)}\qquad\mbox{ and }\qquad
F_{\beta(a,b+1)}(x)=F_{\beta(a,b)}(x)+\frac{x^{a}(1-x)^{b}}{b\mathrm{B}
\left(a,b \right)}
\]
where $B(a,b)=\frac{(a-1)!(b-1)!}{(a+b-1)!}$ is the Beta function.
\end{fact}

For the analysis we use the following theorems regarding the tail of
the $\beta(a,b)$ distribution, when we fix the parameter $b=n+1$ and sum
over parameters $a \ge 1$.

\begin{restatable}{theorem}{fsumboundhalf} \label{thm:F_sum_bound-ym}
For every $n\ge1$ we have $\sum_{i=n+1}^{\infty} F_{\beta(i+1,n+1)}
\left(\frac{1}{2} \right)=O(\sqrt{n})$.
\end{restatable}

\subsection{Notations}
When the bit sequence
$\Gamma= (\gamma_1,\dots, \gamma_T )$ can be inferred from the context, we
use
$O_t$ and $Z_t$ rather than $O_t(\Gamma)$ and $Z_t(\Gamma)$.

We also define the $sign$ function as $ sign(x) = \left\{
\begin{smallmatrix}
    1 & x>0 \\
    0 & x=0 \\
    -1 & x<0
\end{smallmatrix} \right.
$.

For functions $f,g \in \mathbb{R} \rightarrow \mathbb{R}$ we denote
$g=O(f)$ iff there exist $c_1,c_2 \in \mathbb{R}$ such that $g(x)
\le c_1f(x)+c_2$ for every $x \in \mathbb{R}$.

\section{Thompson sampling for bit prediction} \label{sec:thompson-sampling}

The Thompson sampling algorithm requires a prior distribution for
its initialization. Given the observations, it updates the prior
distribution to a posterior distribution. The learner samples the
posterior distribution, and thresholds the sampled value at half (for
bit prediction) or $q$ (for generalized bit prediction).

More specifically. We consider the prior distribution $\beta(1,1)$,
which is a uniform distribution over $[ 0,1 ]$. Note that this prior
is fictitious, and used only to initialize the Thompson sampling
algorithm. At time $t$ the learner samples a value $x_t$ from the
distribution $\beta(O_{t-1}+1,Z_{t-1}+1)$, where $O_{t-1}$ and
$Z_{t-1}$ are the number of observed $1$'s and $0$'s up to time
$t-1$, respectively. At time $t$ the learner predicts
$\hat{\gamma}_t = \mathbb{I} \lbrace x_t > q \rbrace$, where $q$ is
the trade-off parameter of the loss. Then the learner observes the
feedback bit $\gamma_t$ and suffers loss
$\ell^q(\hat{\gamma}_t,\gamma_t)$. The resulting Thompson sampling
algorithm is described in Algorithm~\ref{alg:beta-thompson-step},
and in the analysis we refer to this algorithm as $TS(q)$.

\begin{algorithm}[t]
\caption{Thompson sampling with Beta prior for bit prediction
}\label{alg:beta-thompson-step} \SetKwInOut{Input}{input}
\SetKwInOut{Output}{output} \SetKwInOut{Initialize}{initialize}

\Input{Trade-off parameter $q \in [0,1]$.}
\Initialize{Set $O_0=0,Z_0=0$.} \BlankLine


\For{each time $t$ in $\left[ T \right]$}{
    \emph{Sample $x_t$ from the $\beta \left( O_{t-1} + 1, Z_{t-1} + 1 \right)$ distribution.}

    \emph{Predict bit $\hat{\gamma}_t = \mathbb{I} \lbrace x_t > q \rbrace$.}

    \emph{Observe bit $\gamma_{t}$ and suffer loss $\ell_t=\ell^q(\hat{\gamma}_t , \gamma_t).$}

    \emph{Update $O_{t}=O_{t-1}+\gamma_{t}$ and $Z_{t}=Z_{t-1}+(1-\gamma_{t})$.}
}
\end{algorithm}

In  Section~\ref{sec:swapping-lemma} we prove the ``Swapping
Lemma'', which analyses the effect of a single swap on the regret,
which allows us to identify the sequences with the largest and
smallest regret. In Section \ref{sec:regert-half} we bound the
regret of these sequences, thereby obtaining tight upper and lower
bounds on the regret. Section~\ref{sec:regert-q} addresses the
generalized bit prediction case.


\section{Swapping Lemma} \label{sec:swapping-lemma}

In this section we compare the regret of two bit sequences which
differ by  a single swap. This is an essential building
block in our analysis of the worst case and the best case regret of
the Thompson sampling algorithm.




\noindent{\textbf{Swap operation:}}
Given a bit sequence  $\Gamma= (\gamma_1,\ldots, \gamma_T )$,
performing the swap operation at position $t \in \left[ T \right]$
results in a sequence that swaps $\gamma_t$ and
$\gamma_{t+1}$ in $\Gamma$ and keeps all other bits unchanged.
Formally,
$Swap(\Gamma,t)=(\gamma_{1},\ldots,\gamma_{t-1},\gamma_{t+1},\gamma_{t},\gamma_{t+2},\ldots,\gamma_{T})$.

The swapping lemma that compares the
regret of Thompson sampling, $TS( q )$, on the bit
sequences $\Gamma$ and $Swap(\Gamma,t)$.

To illustrate the swapping lemma consider the case $q=\frac{1}{2}$,
so $\frac{q}{1-q}=1$. If we had more zeros up to position t-1 then having the one earlier increases the regret. If we had more ones up to position t-1 then having zero earlier increases the regret. More precisely, for each $t$ such that $\gamma_t = 0$,
$\gamma_{t+1} =1$ and $O_{t-1} < Z_{t-1}$, swapping $\gamma_t$ and
$\gamma_{t+1}$ increases the regret. Similarly, if $\gamma_t = 1$,
$\gamma_{t+1} =0$ and $O_{t-1} > Z_{t-1}$ then swapping $\gamma_t$
and $\gamma_{t+1}$ increases the regret. In other words, 

\begin{restatable}[Swapping Lemma]{lemma}{swaplemma}\label{lem:swap-lemma}
Fix a bit sequence $\Gamma=\left(\gamma_1,\ldots, \gamma_T \right)
\in \left\lbrace 0,1 \right\rbrace^T$. For every $t$, such that
$\gamma_{t}=0$ and $\gamma_{t+1}=1$, we have
  \[
      Regret_{TS(q)}^q (\Gamma) < Regret_{TS(q)}^q (Swap(\Gamma,t)) \Longleftrightarrow
      \frac{q}{1-q} > \frac{O_{t-1} +1}{Z_{t-1} +1}.
  \]
For every $t$, such that $\gamma_{t}=1$ and $\gamma_{t+1}=0$, we
have
\[
Regret_{TS(q)}^q (\Gamma) < Regret_{TS(q)}^q (Swap(\Gamma,t))
\Longleftrightarrow
      \frac{q}{1-q} < \frac{O_{t-1}+1}{Z_{t-1}+1}.
\]
In addition,
\[
Regret_{TS(q)}^q (\Gamma) = Regret_{TS(q)}^q (Swap(\Gamma,t))
\Longleftrightarrow
      \frac{q}{1-q} = \frac{O_{t-1}+1}{Z_{t-1}+1}.
\]
\end{restatable}

\begin{proof-sketch}
%
We consider the difference between the regret of $TS( q )$ on the
bit sequence  $\Gamma$ and on the bit sequence $Swap(\Gamma,t)$. The two
bit sequences differ only at locations $t$ and $t+1$. Since the
benchmark of a sequence depends only on the total number of zeros
and ones in the sequence, the benchmarks on $\Gamma$ and
$Swap(\Gamma,t)$ are identical, i.e.,
$static^q(\Gamma)=static^q(Swap(\Gamma,t))$. Therefore, the
difference between the regrets is equals to the difference between the losses at
time $t$ and $t+1$.

Consider time $t \in \left[ T \right]$ such that $\gamma_{t}=0$ and
$\gamma_{t+1}=1$.Using the insights above it is easy to show that,
\begin{align*}
R&egret_{TS \left(q\right)}^q\left(\Gamma\right) - Regret_{TS
\left(q\right)}^q\left( Swap(\Gamma,t) \right) \\
&= (1-q)F_{\beta \left(O_{t-1}+1,Z_{t-1}+2\right)}(q) + q F_{\beta
\left(O_{t-1}+2,Z_{t-1}+1\right)}(q) - F_{\beta
\left(O_{t-1}+1,Z_{t-1}+1\right)}(q),
\end{align*}
Using the recurrence relations in Fact~\ref{fact:RBF_properties} we show that,

\begin{align*}
\label{eq:tmp swap eq 1}
\begin{split}
R&egret_{TS(q)}^q (\Gamma) - Regret_{TS(q)}^q(Swap(\Gamma,t))\\
&=
\frac{q^{O_{t-1}+1}(1-q)^{Z_{t-1}+1}}{B\left(O_{t-1}+1,Z_{t-1}+1\right)}
\left( \frac{1-q}{Z_{t-1}+1} - \frac{q}{O_{t-1}+1} \right),
\end{split}
\end{align*}
%
Since
$\frac{q^{O_{t-1}+1}(1-q)^{Z_{t-1}+1}}{B\left(O_{t-1}+1,Z_{t-1}+1\right)}
> 0$, we have
\[
      Regret_{TS(q)}^q \left(\Gamma\right) < Regret_{TS(q)}^q \left(Swap(\Gamma,t)\right) \Longleftrightarrow \\
      \frac{q}{1-q} > \frac{O_{t-1}+1}{Z_{t-1}+1},
\]
and equality holds iff $\frac{q}{1-q} =
\frac{O_{t-1}+1}{Z_{t-1}+1}$.
The second case, where $\gamma_{t}=1$ and $\gamma_{t+1}=0$, is
similar.
\end{proof-sketch}

\section{Regret characterization for
$q=\frac{1}{2}$}\label{sec:regert-half}


 In this section we use the swapping lemma
to characterize the sequences on which $TS ( \frac{1}{2} )$ has  the largest  and smallest regret.  We denote by $k$ the number of zeros in the sequence and
characterize the sequences of worst and best regret for each $k$.
Notice that we may assume that $k\le \frac{T}{2}$ since any sequence $\Gamma$ has the same regret as
the sequence $\Gamma'$ obtained from $\Gamma$ by flipping each bit.
Indeed, $static(\Gamma) = static(\Gamma')$ and the expected loss of
$TS(\frac{1}{2})$ on $\Gamma$ and $\Gamma'$ is the same (by Fact \ref{fact:beta-symmetry}).

\subsection{Worst-case regret} \label{subsec:Worst-case regret half}
Consider bit sequences $\Gamma = ( \gamma_1, \ldots , \gamma_T )$
with $k $ zeros, where $k\le \frac{T}{2}$. We first show that among
these bit sequences the ones of largest regret are of the form
$\lbrace 01,10 \rbrace^k1^{T-2k}$. Then, we prove that the regret of
each of these sequences is $\Theta ( \sqrt{k} )$.

\begin{restatable}{theorem}{opthalf}
\label{thm:opt-half} For any $\Gamma_1,\Gamma_2 \in \lbrace 01,10
\rbrace^k1^{T-2k}$ we have $Regret^{1/2}_{TS( \frac{1}{2} )}(\Gamma_1)=Regret^{1/2}_{TS( \frac{1}{2} )}(\Gamma_2)$. In
addition, for any $\Gamma_3 \notin \lbrace 01,10 \rbrace^k1^{T-2k}$
we have $Regret^{1/2}_{TS( \frac{1}{2} )}(\Gamma_1) >Regret^{1/2}_{TS( \frac{1}{2} )}(\Gamma_3)$.
\end{restatable}


\begin{proof}
Note that for any $i\in[k]$ we have
$O_{2i}(\Gamma_1)=Z_{2i}(\Gamma_1)=i$. By Lemma~\ref{lem:swap-lemma}
this implies that $Regret^{1/2}_{TS( \frac{1}{2} )}(\Gamma_1)=Regret^{1/2}_{TS( \frac{1}{2} )}(Swap(\Gamma_1,i))$. Since
we can transform $\Gamma_1$ to $\Gamma_2$ by a sequence of swap
operations at certain locations $2i$, it follows that
$Regret^{1/2}_{TS( \frac{1}{2} )}(\Gamma_1)=Regret^{1/2}_{TS( \frac{1}{2} )}(\Gamma_2)$. This implies that all the
sequences of the form $\lbrace 01,10 \rbrace^k1^{T-2k}$ have the
same regret.

Let $\Gamma_3=(\gamma_1,\ldots,\gamma_T ) \in \lbrace 0,1 \rbrace^T$
be a bit sequence of length $T$ with $k$ zeros such that $\Gamma_3
\notin \lbrace 01,10 \rbrace^k1^{T-2k}$. We show  that for some
$t\in \left[ T \right]$, the sequence $Swap ( \Gamma_3,t )$ has a
regret larger than $\Gamma_3$.

Since $\Gamma_3 \notin \lbrace 01,10 \rbrace^k1^{T-2k}$,  there is
an index $i \le k-1$ such that either
$\gamma_{2i+1}=\gamma_{2i+2}=1$ or $\gamma_{2i+1}=\gamma_{2i+2}=0$.
Let $i$ to be the smallest such index.
Assume that $\gamma_{2i+1}=\gamma_{2i+2}= 1$. (The case of
$\gamma_{2i+1}=\gamma_{2i+2}=0$ is similar.) It follows that
$O_{2i}=Z_{2i}$ and $O_{2i+1}=Z_{2i+1}+1$.
Let $j > 2i+2$ be the minimal index such that $\gamma_{j} = 0$. Such
an index must exist, since there are $k$ zeros in $\Gamma_3$ and until
index $2i$ there were only $i\le k-1$ zeros. Since $\gamma_{j-1}
=\gamma_{j-2}= 1$ we have $\frac{O_{j-1}}{Z_{j-1}} >
\frac{O_{j-2}}{Z_{j-2}} \ge \frac{O_{2i+1}}{Z_{2i+1}} > 1$. By
Lemma~\ref{lem:swap-lemma}, the sequence $Swap ( \Gamma_3, j-1 )$
has regret higher than $\Gamma_3$, i.e.,
$Regret^{1/2}_{TS( \frac{1}{2} )}(\Gamma_3)<Regret^{1/2}_{TS( \frac{1}{2} )}(Swap(\Gamma_3,t))$.

Since there are finite number of bit sequences of length $T$ with
$k$ zeros, we get that sequences with the largest regret must be of
the form $\lbrace 01,10 \rbrace^k1^{T-2k}$.
\end{proof}

Given the above theorem, to bound the worst case regret of $TS(
\frac{1}{2})$, we can focus on the sequence $W_T^k=\{01\}^k 1^{T-2k}$
and bound $Regret^{1/2}_{TS( \frac{1}{2} )}
(W_T^k)$.

\begin{restatable}{theorem}{zoregrethalf} \label{thm:0-1-regret-half}
For every $T \in \mathbb{N}^+$ and $k \le \frac{T}{2}$ we have,
$Regret^{1/2}_{TS( \frac{1}{2})} (W_T^k)=\Theta(\sqrt{k})$.
\end{restatable}


\begin{proof-sketch}
Let $W_T^k = ( w_1, \ldots,w_T )$, where we have: (1) $w_t =0$ for
$t\in A_1=\{2i-1 \mid i\in [k]\}$, (2) $w_t = 1$ for $t\in A_2=\{2i
\mid i\in[k]\}$, and (3) $w_t=1$ for $t\in A_3=\{i \mid i\geq
2k+1\}$. We  bound the expected number of errors made by $TS (
\frac{1}{2})$ on each of these three subsets. Then, from these
bounds we derive a bound on the loss and the regret. Specifically we prove the following:

\begin{enumerate}
\item
For $t \in A_1$, $Z_t=O_t$ and thus the probability to predict the next bit is $\frac{1}{2}$. Therefore, the expected number of false positive errors in $A_1$ is
\[
\sum\limits _{t=1}^{k} \E \left[\mathbb{I} \{\hat{\gamma}_{t} \neq
w_{t} \}  \mid W_T^k  \right] =\frac{k}{2}.
\]

\item
For $t \in A_2$, $Z_t=O_t+1$ and the difference between the probability to predict 0 and the probability to predict 1 is small and can be bounded. Therefore, the expected number of false negative errors in $A_2$ is
\[
\sum\limits_{i=1}^{k} \E \left[\mathbb{I} \{\hat{\gamma}_{2i} \neq
w_{2i} \}  \mid W_T^k  \right]
= \frac{k}{2} + \Theta ( \sqrt{k} ).
\]

\item
The expected number of false negative in $A_3$ is show to be
\[
  \sum\limits _{t=2k + 1}^{T} \E \left[\mathbb{I} \{\hat{\gamma}_{t} \neq w_{t} \}  \mid W_T^k  \right] =
  \sum\limits _{t=2k + 1}^{T} F_{\beta(t-k+1,k+1)}\left( \frac{1}{2} \right) = O (\sqrt{k} ),
\]
where the last equality follows from Theorem~\ref{thm:F_sum_bound-ym}.
\end{enumerate}
Summing up the errors over $A_1$, $A_2$, and $A_3$, and recalling
that the static prediction makes $\min\{T-k,k\}=k$ errors, we bound
the regret as follows
\[
\sum\limits _{t=1}^{T} \E \left[\mathbb{I} \{\hat{\gamma}_{t} \neq
w_{t} \}  \mid W_T^k   \right] - \min \left\lbrace T-k, k \right\rbrace =
\frac{k}{2} +  \left( \frac{k}{2} + \Theta( \sqrt{k} ) \right) + O
(\sqrt{k} ) -k = \Theta( \sqrt{k} ).
\]
\end{proof-sketch}


Since $k \le \frac{T}{2}$, we have the following corollary.

\begin{corollary} \label{cor:regret-result-bound-half}
For any sequence of length $T$, the regret of $TS ( \frac{1}{2} )$
is at most $O ( \sqrt{T} )$.
\end{corollary}

\begin{remark}
Note that in fact we proved that $Regret^{1/2}_{TS( \frac{1}{2} )} (
\Gamma ) = \Theta ( \sqrt{\min \lbrace O_T(\Gamma), Z_T(\Gamma)
\rbrace })$.
\end{remark}

\subsection{Best-case regret} \label{subsec:Best-case regret half}

In this subsection, we characterize the sequences with the lowest regret and bound them.

\begin{restatable}{theorem}{besthalf} \label{thm:best-given-K-T-half}
The bit sequence with the
lowest regret of length $T$ with $k < \frac{T}{2}$ zeros is $B_T^k=1^{T-k}0^{k}$.
For $k = \frac{T}{2}$, both $1^{T/2}0^{T/2}$ and $0^{T/2}1^{T/2}$
have the lowest regret.
\end{restatable}

We now bound the regret of $B_T^k$.

\begin{restatable}{theorem}{zobesthalf} \label{thm:0-1-best-regret-half}
For every $T\in \mathbb{N}^+$ and $k \le \frac{T}{2}$ we have,
$Regret^{1/2}_{TS( \frac{1}{2} )} (B_T^k)\leq 1$, where $B_T^k=1^{T-k}0^{k}$.
\end{restatable}

\section{Regret characterization for a general $q$} \label{sec:regert-q}
\newcommand{\wcseq}{worst-case sequence }
\newcommand{\wcseqs}{worst-case sequences }
\newcommand{\nwcseq}{worst-case sequence}
\newcommand{\nwcseqs}{worst-case sequences}
\newcommand{\perpart}{head }
\newcommand{\nperpart}{head}
\newcommand{\headpart}{head }
\newcommand{\nheadpart}{head}
\newcommand{\tailpart}{tail }
\newcommand{\ntailpart}{tail}
\newcommand{\fpre}{p }
\newcommand{\nfpre}{p}
To get some intuition regarding this generalization to an arbitrary trade-off parameter $q$ consider the
following simple example.
Assume that  $q=\frac{1}{3}$, and thereby
 $\frac{q}{1-q} = \frac{1}{2}$ and lets construct a  sequence such that we cannot increase the regret by
swapping any pair of consecutive bits.
This sequence   cannot start with a $1$, since if it does then by
the swapping lemma (Lemma~\ref{lem:swap-lemma} we will be able to increase the regret by swapping the first $0$ with the $1$ preceding it.
So we must start with a $0$.
In general we determine  bit $t+1$ by comparing $\frac{O_t+1}{Z_t+1}$ to $\frac{1}{2}$ (i.e., $\frac{q}{1-q}$). If they are equal then the bit in position $t+1$ is either $0$ or
$1$.
If $\frac{O_t+1}{Z_t+1} > \frac{1}{2}$ the bit in position $t+1$ is $0$ since otherwise we will be able to increase the regret by
swapping the first $0$ following position $t+1$ with its preceding $1$.
Similarly, if $\frac{O_t+1}{Z_t+1} < \frac{1}{2}$ the bit in position $t+1$ is $1$ since otherwise we will be able to increase the regret by
swapping the first $1$ following position $t+1$ with its preceding $0$.

It follows that the second bit could be either $0$ or $1$ since $\frac{O_1+1}{Z_1+1} = \frac{q}{1-q} = \frac{1}{2}$. If we have a $0$ at position $2$ then $\frac{O_2+1}{Z_2+1} =  \frac{1}{3} < \frac{1}{2}$ and therefore we must continue
with a $1$ at position $3$. Then we have that $\frac{O_3+1}{Z_3+1} =  \frac{2}{3} > \frac{1}{2}$ so we put $0$ at position $4$,
and we are back in the situation where $\frac{O_4+1}{Z_4+1} = \frac{1}{2}$ so we can choose either $0$ or $1$ at position $5$.
Similarly, if we place a $1$ at position $2$ then we will have to continue with two $0$'s and then we will be free to choose at position $5$ either $0$ or $1$.
It follows that the family of sequences of the form $0\{100,010\}^*x\{1^*,0^*\}$ (where $x$ could be any prefix of $100$ or $010$)
contains all sequences of largest regret. (We will in fact show that they all have the same regret.)

To gain some deeper intuition assume now that
 $q$ is a rational number and $\frac{q}{1-q} = \frac{n_1}{n_2}$ (where $n_1$ and $n_2$ do not have common divisors)  and lets try to construct a sequence that we cannot increase its regret by applying the swapping lemma.
Whenever $\frac{O_t+1}{Z_t+1} = \frac{n_1}{n_2}$ we can choose any bit to position $t+1$.
At this point we have that $n_2(O_t+1) = n_1(Z_t+1)$ and therefore $n_1(Z_t+1)$ is a multiple of $n_2$ and
$n_2(O_t+1)$ is a multiple of $n_1$. Once we choose, say $0$, then we are forced to choose a particular sequence in the following $n_1+n_2-1$ steps, until we will again have that $n_2(O_{t'}+1) = n_1(Z_{t'}+1)$ for
$t' = t + n_1 + n_2$ among these bits $n_2$ would be zeros and $n_1$ would be ones so $Z_{t'} = Z_t + n_2$
$O_{t'} = O_t + n_1$.

The structure of this section is similar to the structure of Section~\ref{sec:regert-half}. First, we characterize the bit sequences of
largest regret. Then, we bound the regret of these sequences.


\subsection{Worst-case sequences} \label{subsec:wc-seq}
Consider the following function that maps a bit-sequence to a set of bits
\begin{equation}
\begin{split}
\forall \Phi \in \left\lbrace 0,1 \right\rbrace^*: H^{q} \left( \Phi \right) = \left\{
  \begin{matrix}
    \{ 0 \} & \frac{O( \Phi ) + 1}{Z( \Phi ) + 1} > \frac{q}{1-q} \\
    \{ 1 \} & \frac{O( \Phi ) + 1}{Z( \Phi ) + 1} < \frac{q}{1-q} \\
    \{ 0,1 \} & \frac{O( \Phi ) + 1}{Z( \Phi ) + 1} = \frac{q}{1-q}
  \end{matrix}
\right. ,
\end{split}
\end{equation}
where $O(\Phi)$ is the total number of $1$s in $\Phi$ and $Z(\Phi)$ is the total number of $0$s in $\Phi$.

For every  sequence $\Gamma = (\gamma_1,\ldots,\gamma_T) \in \{0,1\}^T$ we define  $\nfpre(\Gamma)$ to be the largest index $t$ s.t. $\forall i \in [ t ]: \gamma_i \in H^{q} ( \Gamma_{1:i-1} )$, where $ \Gamma_{1:n}=(\gamma_1,\ldots,\gamma_n)$. We call a bit sequence $\Gamma=(\gamma_1,\ldots,\gamma_T )$ a \textit{\nwcseq} if $\gamma_{\nfpre(\Gamma)+1}=\ldots=\gamma_T$. We define the subsequence $(\gamma_1,\ldots,\gamma_{\nfpre(\Gamma)})$ as the \textit{\perpart} of $\Gamma$ and denote it
$\perpart(\Gamma)$ and the subsequence $(\gamma_{\nfpre(\Gamma)+1},\ldots,\gamma_T)$ as the \textit{\ntailpart}
of $\Gamma$ and denote it
$\ntailpart(\Gamma)$.

For start, we characterize the \tailpart of a \nwcseq.

\begin{restatable}{theorem}{tailfill} \label{cor:tail-fill}
Let $\Gamma$ be a \nwcseq. If $Z_T \le (1-q)T - q$ then the $\ntailpart(\Gamma)$  is filled with ones. Otherwise, the $\ntailpart(\Gamma)$  is filled with zeros.
\end{restatable}

\subsection{Worst-case regret} \label{subsec:wc-reg-q}
In this subsection we prove that all the \wcseqs have the largest regret and prove an upper bound on this regret.

\begin{restatable}{theorem}{worstkt} \label{thm:worst-given-K-T}
Let $\Gamma \in \lbrace 0,1 \rbrace^T$, s.t.
$\Gamma$ is not a \nwcseq. Then, there exists
$t\in[T]$ such that $Regret^q_{TS(q)}(\Gamma)<Regret^q_{TS(q)}(Swap(\Gamma,t))$.
\end{restatable}

\begin{proof}
Let $i=\nfpre(\Gamma)+1$.
Since $\Gamma$ is not a \nwcseq,
there is an index $j>i$ such that $\gamma_j\not= \gamma_i$ (since, from Theorem~\ref{cor:tail-fill}, $\ntailpart(\Gamma)$ contains both $0$'s and $1$'s).
Assume $j$ is the smallest index with this property.

\noindent
{\bf Case 1} Assume $\gamma_i=0$ and $\gamma_j=1$. Since $\gamma_i \notin H^q(\Gamma_{1:i-1})$ we have $\frac{O_{i-1} (\Gamma) + 1}{Z_{i-1}(\Gamma) + 1} < \frac{q}{1-q}$. From the definition of $j$ follows that $\gamma_{i}= \gamma_{i+1} = \ldots =\gamma_{j-1} = 0$ and thus $\frac{O_{j-2}(\Gamma) + 1}{Z_{j-2}(\Gamma) + 1} \le \frac{O_{i-1}(\Gamma) + 1}{Z_{i-1}(\Gamma) + 1} < \frac{q}{1-q}$. By Lemma \ref{lem:swap-lemma}, the sequence $Swap ( \Gamma, j-1 )$ has a regret larger than $\Gamma$.

\noindent
{\bf Case 2} Assume $\gamma_i=1$ and $\gamma_j=0$. Since $\gamma_i \notin H^q(\Gamma_{1:i-1})$ we have $\frac{O_{i-1} (\Gamma) + 1}{Z_{i-1}(\Gamma) + 1} > \frac{q}{1-q}$. From the definition of $j$ follows that $\gamma_{i}= \gamma_{i+1} = \ldots =\gamma_{j-1} = 1$ and thus $\frac{O_{j-2}(\Gamma) + 1}{Z_{j-2}(\Gamma)  + 1} \ge \frac{O_{i-1}(\Gamma)  + 1}{Z_{i-1}(\Gamma )+ 1} > \frac{q}{1-q}$. By Lemma \ref{lem:swap-lemma}, the sequence $Swap ( \Gamma, j-1 )$ has a regret larger than $\Gamma$.
\end{proof}

Theorem \ref{thm:worst-given-K-T} implies that any
sequence of largest regret is a \nwcseq.
Next we prove that all \wcseqs of length $T$ with $k$ zeros  have the same regret.

\begin{restatable}{lemma}{sameregretq} \label{lem:same-regretq}
All the \wcseqs of length $T$ with $k$ zeros have the same regret.
\end{restatable}

Let $W_T^k=(w_1,\ldots,w_T) \in \{ 0,1 \}^T$
be a \wcseq with $k$ zeros such that for all $t\le \nfpre(W_T^k)$
with $\frac{O_{t-1}+1}{Z_{t-1}+1} = \frac{q}{1-q}$ we have
$\gamma_t = 0$.
 Since
 by Lemma~\ref{lem:same-regretq} all the \wcseqs with the same number of zeros have the same regret, we can focus on
bounding the regret of $W_T^k$.

\begin{restatable}{theorem}{zoregretq} \label{thm:0-1-regret}
For every $T\in \mathbb{N}^+$, $q \in \left[0,\frac{1}{2}\right]$
and $k$ zeros we have
\begin{equation*}
\begin{split}
Regret^q_{TS(q)} (W_T^k) = \left\{ \begin{matrix}
O ( \sqrt{qk} ) & k \le (1-q)T - q \\
O ( \sqrt{(1-q)(T-k)} ) & k > (1-q)T - q
\end{matrix} \right. .
\end{split}
\end{equation*}
\end{restatable}

The regret bounds for $q\in[\frac{1}{2},1]$ are derived from the
Theorem~\ref{thm:0-1-regret} using the following lemma.

\begin{restatable}{lemma}{finalregretboundq} \label{cor:final-regret-bound}
For every bit sequence $\Gamma =( \gamma_1,\ldots,\gamma_T )$ define
$\bar{\Gamma} =( 1-\gamma_1,\ldots,1-\gamma_T )$. Then,
$Regret^q_{TS(q)} \left( \Gamma \right)=Regret^{1-q}_{TS(1-q)}$
\end{restatable}

The following theorem derives the worst-case sequences regret bound
for general $q$.

\begin{restatable}{theorem}{worstqlarge} \label{cor:worst-q-large}
For any observation sequence of length $T$, the regret of $TS ( q )$ is $O \left(\sqrt{q(1-q)T}\right)$.
\end{restatable}

\subsection{Best-case regret bound} \label{subsec:best-reg-q}

We do not characterize the exact best-case regret
sequences\footnote{Finding the best-case sequence characterization
for a general trade-off parameter $q$ is harder than the previous
cases.  With the tools we presented, it is difficult even to compare
the regrets of the bit sequences $10^k$ and $0^k1$ for $k \in
\mathbb{N}$.}, but only show that there are sequence with regret at
most $1$.

\begin{restatable}{theorem}{zobestregretq} \label{thm:0-1-best-regret-q}
For every $q\in (0,1)$ and $m,n \in \mathbb{N}$, if $qm\leq (1-q)n$,
then $Regret^{q}_{TS( q )} (1^n0^m)\leq 1$ and otherwise
$Regret^{q}_{TS( q )} (0^m1^n)\leq 1$.
\end{restatable}

\acks{This work was supported in part by the Yandex Initiative in Machine Learning and by a grant from the Israel Science Foundation (ISF).}

\newpage
\bibliographystyle{unsrt}
\bibliography{biblist}

\newpage
\appendix

\section{Beta and Binomial concentration bounds}
The following identities are well known
(see, for example, \cite{agrawal2017near}, Fact 3 and \cite{NIST:DLMF},
Eq.~8.17.4). 

The first relates the CDFs of the Beta and the Binomial distributions. The second is a property of the Beta distribution. 

\begin{fact}\label{fact:beta-bin-correlation}
For $a,b \in \mathbb{N^+}$ and $p \in [0,1]$ we have $F_{\beta(a,b)}(p)  =1-F_{Bin(a+b-1,p)}(a-1)$.
\end{fact}

\begin{fact}\label{fact:beta-pow}
For $a \in \mathbb{N^+}$ and $p \in [0,1]$ we have $F_{\beta(a,1)}(p)=p^a$.
\end{fact}

Next, we present concentration bounds and inequalities that we need for
our proofs.

\begin{fact}{(Gaussian Half CDF)}\label{fact:gauss-half-cdf}

Let $\sigma \in \mathbb{R^+}$. Then
$\frac{1}{\sqrt{2\pi \sigma^2}}\int\limits_{0}^{\infty} e^{-\frac{x^2}{2\sigma^2}} dx=\frac{1}{2}$.
\end{fact}

\begin{fact}{(Multiplicative Chernoff bound)} \cite{mitzenmacher2005probability}
\label{fact:Chernoff}

Let $X_{1},...,X_{n}$ be random variables with values of $\{0,1\}$
such that $\E[X_{t}|X_{1},...,X_{t-1}]=\mu$. Let $S_{n}=\sum\limits
_{i=1}^{n}X_{i}$. 
\begin{enumerate}
\item For $1 \ge a \ge 0$, $\Pr \left( S_{n} \ge (1+a) n \mu \right) \le e^{-\frac{a^2 n \mu}{3}}$.
\item For $a \ge 1$, $\Pr \left( S_{n} \ge (1+a) n \mu \right) \le e^{-\frac{a n \mu}{3}}$.
\end{enumerate}
\end{fact}

\begin{fact}{(Chernoff-Hoeffding)} \cite{hoeffding1963probability}
\label{fact:Chernoff-Hoffding}

Let $X_{1},...,X_{n}$ be random variables with common range $[0,1]$
such that $\E[X_{t}|X_{1},...,X_{t-1}]=\mu$. Let $S_{n}=\sum\limits
_{i=1}^{n}X_{i}$. 
\begin{enumerate}
\item For all $a \ge 0$,
$\Pr \left( \lvert S_{n} - n\mu \rvert \ge a \right) \le
2e^{-\frac{2a^{2}}{n}}$.
\item
For $\mu \ge \frac{1}{2}$ and $a \ge 0$,
$\Pr \left( S_{n} > n\mu + a \right) \le
e^{-\frac{a^2}{2n\mu(1-\mu)}}$.
\end{enumerate}
\end{fact}
\section{Proof of bounds on sums of Beta CDFs (Theorems \ref{thm:F_sum_bound-ym} and \ref{thm:F_sum_bound-generalized})}

We present two bounds for sums of Beta CDFs.
In the first subsection we prove a simple version of our bound, which appears Theorem~\ref{thm:F_sum_bound-ym}.
In the second subsection we expend the result to a general $q\in (0,1)$.
\subsection{Proof of Theorem~\ref{thm:F_sum_bound-ym}} \label{subsec:app-thm-3}
The proof is divided into two parts. First we prove a bound on a series of exponents and then use Hoeffding bound to show that the exponent series is an upper bound for the sum of beta-distribution CDFs appears in Theorem~\ref{thm:F_sum_bound-ym}.
\begin{lemma} \label{lem:e_sum_bound}
For every $n\ge1$, $\sum\limits
_{i=n+1}^{\infty}e^{-\frac{(i-(n+1))^{2}}{2(i+n+1)}}=\Theta(\sqrt{n})$.
\end{lemma}
\begin{proof}
Let $j = i - (n+1)$, then
\begin{equation} \label{eq:some-tmp-eq}
\sum\limits _{i=n+1}^{\infty}e^{-\frac{(i-(n+1))^{2}}{2(i+n+1)}} =
\sum\limits _{j=0}^{\infty}e^{-\frac{j^{2}}{2 \left(j+2 \left(
n+1\right) \right)}}.
\end{equation}

We bound from below and above the exponents.
For the upper bound we use the fact that $j\geq 0$ and for lower
bounding the exponent we consider two cases: (a) $j > 2(n+1)$ and
(b) $2(n+1) \ge j \ge 0$. We have,
\[
\frac{j^2}{4(n+1)} \ge \frac{j^{2}}{2 \left(j+2 \left( n+1\right)
\right)} \ge \left\{
                            \begin{array}{ll}
                                \frac{j^{2}}{8 \left(n+1 \right)} & 2(n+1) \ge j \ge 0 \\
                                \frac{j}{4} & j > 2(n+1) \\
                            \end{array}
                        \right. .
\]

We bound the sum (\ref{eq:some-tmp-eq}) from below using Fact~\ref{fact:gauss-half-cdf}, where $\sigma^2=2(n+1)$, as follows
\[
\sum\limits _{j=0}^{\infty}e^{-\frac{j^{2}}{2 \left(j+2 \left(
n+1\right) \right)}} \geq
\sum\limits_{j=0}^{\infty}e^{-\frac{j^2}{4(n+1)}} \geq \sqrt{4\pi (n+1)}\frac{1}{\sqrt{4\pi (n+1)}}\int\limits_{0}^{\infty}e^{-\frac{x^2}{4(n+1)}} dx=\sqrt{\pi (n+1)}.
\]

For upper bounding Eq. (\ref{eq:some-tmp-eq}) we have,
\begin{equation} \label{eq:foo-bar}
\sum\limits_{j=0}^{\infty}e^{-\frac{j^{2}}{2 \left(j+2 \left(
n+1\right) \right)}} \leq
\sum\limits_{j=0}^{2(n+1)}e^{-\frac{j^2}{8(n+1)}} +
\sum\limits_{j=2(n+1)}^{\infty}e^{-\frac{j}{4}}.
\end{equation}
The first sum of the right side of Eq. (\ref{eq:foo-bar}) is bounded, by using Fact~\ref{fact:gauss-half-cdf} with $\sigma^2=4(n+1)$, as follows
\[
\sum\limits_{j=0}^{2(n+1)}e^{-\frac{j^2}{8(n+1)}} \le 1 +
\int\limits_{0}^{2(n+1)}e^{-\frac{x^2}{8(n+1)}} dx \le 1+ \sqrt{2
\pi \left( n+1 \right)}.
\]
The second sum of the right hand side of Eq. (\ref{eq:foo-bar}) is
an exponential sum and bounded as follows,
\[
\sum\limits_{j=2(n+1)}^{\infty}e^{-\frac{j}{4}} = \frac{1}{{1 -
e^{-\frac{1}{4}}}} -  \frac{1 - \left( e^{-\frac{1}{4}}
\right)^{2n+3}}{1 - e^{-\frac{1}{4}}} \le \frac{1}{{1 -
e^{-\frac{1}{4}}}}.
\]
By combining the previous inequalities and Eq. (\ref{eq:foo-bar}) we
get $\sum\limits
_{i=n+1}^{\infty}e^{-\frac{(i-(n+1))^{2}}{2(i+n+1)}}=\Theta(\sqrt{n})$.
\end{proof}

\fsumboundhalf*

\begin{proof}
Using Fact~\ref{fact:beta-bin-correlation}
\begin{align*}
\sum\limits
_{i=n+1}^{\infty}F_{\beta(i+1,n+1)}\left(\frac{1}{2}\right) &=
\sum\limits _{i=n+1}^{\infty}\left(1-F_{Bin \left(i+n+1,\frac{1}{2} \right)}(i)\right) \\
&= \sum\limits _{i=n+1}^{\infty}\left(1-\Pr_{x_{j}\sim Ber\left(\frac{1}{2}\right)}\left(\sum\limits _{j=1}^{i+n+1}x_{j}\le i\right)\right) \\
&= \sum\limits _{i=n+1}^{\infty}\Pr_{x_{j}\sim
Ber\left(\frac{1}{2}\right)}\left(\sum\limits _{j=1}^{i+n+1}x_{j} -
\frac{i+n+1}{2} \geq \frac{i-(n+1)}{2}\right).
\end{align*}

Note that $\frac{i-(n+1)}{2} \ge 0$ when $i \ge n+1$, therefore we
can use the Chernoff-Hoffding bound
(Fact~\ref{fact:Chernoff-Hoffding}.1) to achieve
\[
\sum\limits
_{i=n+1}^{\infty}F_{\beta(i+1,n+1)}\left(\frac{1}{2}\right) \le 2
\sum\limits _{i=n+1}^{\infty}e^{-\frac{(i-(n+1))^{2}}{2(i+n+1)}} =
\Theta(\sqrt{n}).
\]
where the last equality follows from Lemma~\ref{lem:e_sum_bound}.
\end{proof}

\subsection{Proof of Theorem~\ref{thm:F_sum_bound-generalized}}

The following subsection generalizes the proof of Theorem~\ref{thm:F_sum_bound-ym}, as presented in Appendix~\ref{subsec:app-thm-3}.
We divide the generalized theorem version proof into two parts similarly to Appendix~\ref{subsec:app-thm-3}.
\begin{lemma}
\label{lem:e_sum_bound_q}For every $n \in \mathbb{N}^+$, $a>0$ and $p\in(0,1)$ we have
\begin{enumerate}
\item
$\sum\limits _{i=\ceil*{\frac{p}{1-p}(n+1)}+1}^{\infty} e^{-\frac{((1-p)i - p(n+1))^2}{a(i+n+1)}} \le
\frac{\sqrt{\pi a(n+1)}}{\sqrt{2}(1-p)^{3/2}}+\frac{2a}{(1-p)^2}e^{-\frac{1-p}{2a}(n+1)}$,
\item
$\sum\limits _{i=\floor*{\frac{2p(n+1)}{1-2p}} + 1}^{\infty} e^{-\frac{(1-p)i - p(n+1)}{a}} \le
1+\frac{a}{1-p} e^{-\frac{p(n+1)}{a(1-2p)}}$ .
\end{enumerate}
\end{lemma}
\begin{proof}

\noindent
\textbf{1.}
We bound the sum as follows
\begin{equation*}
\sum\limits _{i=\ceil*{\frac{p}{1-p}(n+1)}+1}^{\infty} e^{-\frac{((1-p)i - p(n+1))^2}{a(i+n+1)}}
\le \int\limits_{\frac{p}{1-p}(n+1)}^{\infty}e^{-\frac{((1-p)x-p(n+1))^{2}}{a(x+n+1)}}dx.
\end{equation*}
Using a substitution of $y=(1-p)x-p(n+1)$,
\begin{equation} \label{eq:exp-1-1}
\int\limits_{\frac{p}{1-p}(n+1)}^{\infty}e^{-\frac{((1-p)x-p(n+1))^{2}}{a(x+n+1)}}dx
\le \frac{1}{1-p}\int\limits_{0}^{\infty}e^{-\frac{y^2}{a(\frac{y+p(n+1)}{1-p}+n+1)}}dy = \frac{1}{1-p}\int\limits_{0}^{\infty}e^{-\frac{1-p}{a(y+n+1)}y^2}dy.
\end{equation}

We bound the exponent from below by considering two cases $y>n+1$ and
$n+1 \ge y \ge 0$. We have,
\begin{equation*}
\frac{1-p}{a(y+n+1)}y^2 \ge \left\{
                            \begin{array}{ll}
                                \frac{1-p}{2a(n+1)}y^{2} & n+1 \ge y \ge 0 \\
                                \frac{1-p}{2a}y & y > n+1 \\
                            \end{array}
                        \right. .
\end{equation*}
Hence, we have
\begin{equation}\label{eq:aaaaa-1-2}
\int\limits_{0}^{\infty}e^{-\frac{1-p}{a(y+n+1)}y^2}dy \le \int\limits_{0}^{n+1}e^{-\frac{1-p}{2a(n+1)}y^2}dy + \int\limits_{n+1}^{\infty}e^{-\frac{1-p}{2a}y}dy .
\end{equation}

We bound the first integral of Eq. (\ref{eq:aaaaa-1-2}) using Fact~\ref{fact:gauss-half-cdf}, where $\sigma^2=\frac{a(n+1)}{1-p}$, as follows
\begin{equation} \label{eq:exp-1-3}
\int\limits_{0}^{n+1}e^{-\frac{1-p}{2a(n+1)}y^2}dy \le \sqrt{\frac{2\pi a(n+1)}{1-p}}\sqrt{\frac{1-p}{2\pi a(n+1)}}\int\limits_{0}^{\infty}e^{-\frac{1-p}{2a(n+1)}y^2}dy = \sqrt{\frac{\pi a(n+1)}{2(1-p)}}.
\end{equation}
The second integral in Eq. (\ref{eq:aaaaa-1-2}) equals
\begin{equation} \label{eq:exp-1-4}
\int\limits_{n+1}^{\infty}e^{-\frac{1-p}{2a}y}dy  = \frac{2a}{1-p}e^{-\frac{1-p}{2a}(n+1)}.
\end{equation}
Combining Eq. (\ref{eq:exp-1-1} - \ref{eq:exp-1-4}) we have
\begin{equation*}
\sum\limits _{i=\ceil*{\frac{p}{1-p}(n+1)}+1}^{\infty} e^{-\frac{((1-p)i - p(n+1))^2}{a(i+n+1)}} \le
 \frac{\sqrt{\pi a(n+1)}}{\sqrt{2}(1-p)^{3/2}}+\frac{2a}{(1-p)^2}e^{-\frac{1-p}{2a}(n+1)}.
\end{equation*}

\noindent
\textbf{2.} We bound the sum as follows
\begin{equation*}
\sum\limits_{i=\floor*{\frac{2p(n+1)}{1-2p}} + 1}^{\infty} e^{-\frac{(1-p)i - p(n+1)}{a}} \le 1+ \int\limits_{\frac{2p(n+1)}{1-2p}}^{\infty} e^{-\frac{(1-p)x-p(n+1)}{a}}dx.
\end{equation*}
Using a substitution of $y=(1-p)x-p(n+1)$,
\begin{equation*}
1+ \int\limits_{\frac{2p(n+1)}{1-2p}}^{\infty} e^{-\frac{(1-p)x-p(n+1)}{a}}dx \le 1+ \frac{1}{1-p}\int\limits_{\frac{p(n+1)}{1-2p}}^{\infty} e^{-\frac{y}{a}}dy = 1+\frac{a}{1-p} e^{-\frac{p(n+1)}{a(1-2p)}}.
\end{equation*}
\end{proof}

\begin{theorem} \label{thm:F_sum_bound-generalized}
For every $n\ge1$ and $p\in (0,1)$ we have
\[
\sum\limits _{i=\floor*{\frac{p}{1-p}n}+1}^{\infty}F_{\beta(i+1,n+1)}(p)=
\left\{
\begin{matrix}
2\sqrt{3\pi p(n+1)} + O(1) & p \le \frac{1}{2} \\
1+\frac{p}{1-p}+\frac{\sqrt{\pi p(n+1)}}{1-p}+\frac{4p}{1-p}e^{-\frac{1}{4p}(n+1)} & p \ge \frac{1}{2}
\end{matrix}
\right. .
\]
\end{theorem}

\begin{proof}
Using Fact~\ref{fact:beta-bin-correlation}
\begin{align} \label{eq:F_sum_q_1_tmp}
\sum\limits _{i=\floor*{\frac{p}{1-p}n}+1}^{\infty}F_{\beta(i+1,n+1)}(p) &=
\sum\limits _{i=\floor*{\frac{p}{1-p}n}+1}^{\infty}\left(1-F_{Bin(i+n+1,p)}(i)\right)   \\
&= \sum\limits _{i=\floor*{\frac{p}{1-p}n}+1}^{\infty}\left(1-\Pr_{X_{j}\sim Ber(p)}\left(\sum\limits _{j=1}^{i+n+1}X_{j}\le i\right)\right) \nonumber \\
&= \sum\limits _{i=\floor*{\frac{p}{1-p}n}+1}^{\infty}\Pr_{X_{j}\sim Ber(p)}\left(\sum\limits _{j=1}^{i+n+1}X_{j} > i\right). \nonumber
\end{align}
Let $N_i=i+n+1$ and $r_i=(1-p)i - p(n+1)$. We have $i=pN_i + r_i$ and therefore, we rewrite Eq. (\ref{eq:F_sum_q_1_tmp}) as
\begin{equation} \label{eq:F_sum_q_1}
\sum\limits_{i=\floor*{\frac{p}{1-p}n}+1}^{\infty}F_{\beta(i+1,n+1)}(p) = \sum\limits_{i=\floor*{\frac{p}{1-p}n}+1}^{\infty}\Pr_{X_{j}\sim Ber(p)}\left(\sum\limits_{j=1}^{N_i}X_{j} > pN_i + r_i\right).
\end{equation}

\noindent
\textbf{1.}
First, we focus on the case of $p \le \frac{1}{2}$.

Consider $\frac{r_i}{pN_i}$ and notice that $1 > \frac{r_i}{pN_i} \ge 0$ when $1 > \frac{(1-p)i - p(n+1)}{p(i+n+1)} \ge 0$, which is equivalent to $\frac{2p}{1-2p}(n+1) > i \ge \frac{p}{1-p}(n+1)$.
Also, we note that $\E_{X_{j}\sim Ber(p)}[\sum\limits _{j=1}^{N_i}X_{j}]=pN_i$. Using Chernoff bound (Fact~\ref{fact:Chernoff}.1) and Lemma~\ref{lem:e_sum_bound_q}.1, with $a=3p$, we have
\begin{align} \label{eq:F_sum_q_2}
\sum\limits_{i=\ceil*{\frac{p}{1-p}(n+1)}+1}^{\floor*{\frac{2p}{1-2p}(n+1)}}&\Pr_{X_{j}\sim Ber(p)}\left(\sum\limits _{j=1}^{N_i}X_{j} > pN_i + r_i\right) \le \sum\limits_{i=\ceil*{\frac{p}{1-p}(n+1)}+1}^{\floor*{\frac{2p}{1-2p}(n+1)}} e^{-\frac{r_i^2}{3pN_i}}  \\ &\le \sum\limits_{i=\ceil*{\frac{p}{1-p}(n+1)}+1}^{\infty} e^{-\frac{((1-p)i - p(n+1))^2}{3p(i+n+1)}}
 \le \frac{\sqrt{3\pi p(n+1)}}{\sqrt{2}(1-p)^{3/2}}+\frac{6p}{(1-p)^2}e^{-\frac{1-p}{6p}(n+1)}. \nonumber
\end{align}
When $i > \frac{2p}{1-2p}(n+1)$ we use the second form of Chernoff bound (Fact~\ref{fact:Chernoff}.2), followed by Lemma~\ref{lem:e_sum_bound_q}.2, with $a=3$, to have
\begin{align} \label{eq:F_sum_q_3}
\sum\limits_{i=\floor*{\frac{2p}{1-2p}(n+1)}+1}^{\infty}&\Pr_{X_{j}\sim Ber(p)}\left(\sum\limits_{j=1}^{N_i}X_{j} > pN_i + r_i\right) \le \sum\limits_{i=\floor*{\frac{2p}{1-2p}(n+1)}+1}^{\infty} e^{-\frac{r_i}{3}}
\\ &= \sum\limits_{i=\floor*{\frac{2p}{1-2p}(n+1)}+1}^{\infty} e^{-\frac{(1-p)i - p(n+1)}{3}} \le
1+\frac{3}{1-p} e^{-\frac{p(n+1)}{3(1-2p)}}. \nonumber
\end{align}
When $\frac{p}{1-p}(n+1)>i$ we can assume worst-case to get
\begin{align} \label{eq:F_sum_q_4}
\sum\limits_{i=\floor*{\frac{p}{1-p}n}+1}^{\ceil*{\frac{p}{1-p}(n+1)}}\Pr_{X_{j}\sim
Ber(p)}\left(\sum\limits_{j=1}^{N_i}X_{j} > pN_i + r_i\right) \le 1+
\frac{p}{1-p}.
\end{align}
By substituting Eq. (\ref{eq:F_sum_q_2}-\ref{eq:F_sum_q_4}) in Eq. (\ref{eq:F_sum_q_1}) we have
\begin{align*}
\sum\limits_{i=\floor*{\frac{p}{1-p}n}+1}^{\infty}F_{\beta(i+1,n+1)}(p)
 \le 2&+ \frac{p}{1-p} + \frac{\sqrt{3\pi p(n+1)}}{\sqrt{2}(1-p)^{3/2}}
 \\&+\frac{6p}{(1-p)^2}e^{-\frac{1-p}{6p}(n+1)} + \frac{3}{1-p} e^{-\frac{p(n+1)}{3(1-2p)}}.
\end{align*}
Since $p \le \frac{1}{2}$, we have $\frac{1}{2}\le 1-p$, thus
\begin{align*}
\sum\limits_{i=\floor*{\frac{p}{1-p}n}+1}^{\infty}F_{\beta(i+1,n+1)}(p) = 2\sqrt{3\pi p(n+1)} + O(1).
\end{align*}

\noindent
\textbf{2.}
Now, consider $p \ge \frac{1}{2}$.
Assume $i \ge \frac{p}{1-p}(n+1)$ and therefore $r_i = (1-p)i - p(n+1) \ge pn+p-pn-p=0$.
Using Hoeffding bound (Fact~\ref{fact:Chernoff-Hoffding}.2) we get that
\begin{equation*}
\Pr_{X_{j}\sim Ber(p)}\left(\sum\limits_{j=1}^{N_i}X_{j} > pN_i + r_i\right) \le e^{-\frac{r_i^{2}}{2p(1-p)N_i}}.
\end{equation*}
Thus, by using Lemma \ref{lem:e_sum_bound_q}.1, with $a=2p(1-p)$, we have
\begin{align} \label{eq:cher-hoeff2}
\sum\limits_{i=\ceil*{\frac{p}{1-p}(n+1)}+1}^{\infty}&\Pr_{X_{j}\sim Ber(p)}\left(\sum\limits_{j=1}^{N_i}X_{j} > pN_i + r_i\right) \le \sum\limits_{i=\ceil*{\frac{p}{1-p}(n+1)}+1}^{\infty} e^{-\frac{r_i^{2}}{2p(1-p)N_i}}
\nonumber \\&\le \frac{\sqrt{\pi p(n+1)}}{1-p}+\frac{4p}{(1-p)}e^{-\frac{1}{4p}(n+1)}.
\end{align}

For $i \le \frac{p}{1-p}(n+1)$ we assume the worst-case bound to get
\begin{equation} \label{eq:cher-hoeff3}
\sum\limits_{i=\floor*{\frac{p}{1-p}n}+1}^{\ceil*{\frac{p}{1-p}(n+1)}}\Pr_{X_{j}\sim Ber(p)}\left(\sum\limits_{j=1}^{N_i}X_{j} > pN_i + r_i\right) \le 1+\frac{p}{1-p}.
\end{equation}

By substituting Eq. (\ref{eq:cher-hoeff2}, \ref{eq:cher-hoeff3}) in Eq. (\ref{eq:F_sum_q_1}) and using Lemma \ref{lem:e_sum_bound_q}.1, with $a=2p(1-p)$, to have
\begin{align*}
\sum\limits_{i=\floor*{\frac{p}{1-p}n}+1}^{\infty}F_{\beta(i+1,n+1)}(p) \le
1+\frac{p}{1-p}+\frac{\sqrt{\pi p(n+1)}}{1-p}+\frac{4p}{(1-p)}e^{-\frac{1}{4p}(n+1)}.
\end{align*}
\end{proof}

\section{Proof of the Swapping Lemma (Lemma~\ref{lem:swap-lemma})}
We start with the following preliminary lemma that states the
probability of an error for $TS( q )$ given a history.
\begin{lemma}
\label{fact:The-expected-reward} Fix a bit sequence
$\Gamma=(\gamma_1,\ldots, \gamma_T ) \in \lbrace 0,1 \rbrace^T$. For
any $t\in[T]$ we have,
\[
\Pr [ \hat{\gamma}_t \neq \gamma_t \mid\Gamma]=E [\mathbb{I}
\{\hat{\gamma}_t \neq \gamma_t\} \vert \Gamma ] =
            \left\{
                \begin{matrix}
                    1-F_{\beta(O_{t-1}+1,Z_{t-1}+1 )}(q) & \gamma_t=0 \\
                    F_{\beta(O_{t-1}+1,Z_{t-1}+1 )}(q) & \gamma_t=1
                \end{matrix}
             \right.
\]
\end{lemma}
\begin{proof}
At time $t$, algorithm $TS ( q )$ samples  $x_t
\sim \beta \left(O_{t-1}+1,Z_{t-1}+1 \right)$, and predicts
$\hat{\gamma}_t = 1$ if  $x_t > q$ and $\hat{\gamma}_t = 0$ if
$x_t \leq q$.
Thus, for the case of $\gamma_t = 0$,
\[
\Pr \left( \hat{\gamma}_t \neq \gamma_t=0 \right) = \Pr \left( x_t > q
\right) = 1 - F_{\beta\left(O_{t-1}+1,Z_{t-1}+1 \right)}(q),
\]
and for the case of $\gamma_t = 1$,
\[
\Pr \left( \hat{\gamma}_t \neq \gamma_t =1\right) = \Pr \left( x_t \le
q \right) = F_{\beta\left(O_{t-1}+1,Z_{t-1}+1 \right)}(q).
\]
\end{proof}

Now we can prove the Swapping Lemma, which compares the regret of two
sequences that differ by a single swap operation.

\swaplemma*
\begin{proof}
%
We consider the difference between the regret of $TS( q )$
on the bit sequence  $\Gamma$ and the bit sequence $Swap(\Gamma,t)$.
The two bit sequences differ only at locations $t$ and $t+1$. Since
the benchmark of a sequence depends only on the total number of
zeros and ones in the sequence, the benchmarks on $\Gamma$ and
$Swap(\Gamma,t)$ are identical, i.e., $statis^q(\Gamma)=static^q(Swap(\Gamma,t))$.
Therefore, the difference between
the regrets is equals to the loss difference at time $t$ and $t+1$.

Consider time $t \in \left[ T \right]$ such that $\gamma_{t}=0$ and
$\gamma_{t+1}=1$.We have,
\begin{align*}
R&egret_{TS \left(q\right)}^q\left(\Gamma\right) - Regret_{TS
\left(q\right)}^q\left( Swap(\Gamma,t) \right) \\
&= \sum\limits_{t=1}^{T} E \left[ \ell^q(\hat{\gamma}_t, \gamma_t)
\mid \Gamma \right] - \sum\limits_{t=1}^{T} E \left[
\ell^q(\hat{\gamma}_t, \gamma_t) \mid Swap(\Gamma,t) \right]
 \\
&= E \left[ \ell^q(\hat{\gamma}_t, \gamma_t) \mid \Gamma \right]+E
\left[ \ell^q(\hat{\gamma}_{t+1}, \gamma_{t+1}) \mid \Gamma \right]
\\ &\phantom{{}=}- \left(E \left[ \ell^q(\hat{\gamma}_t, \gamma_t)
\mid Swap(\Gamma,t) \right]+E \left[ \ell^q(\hat{\gamma}_{t+1},
\gamma_{t+1}) \mid Swap(\Gamma,t) \right] \right)
\\
&= E \left[ \ell^q(\hat{\gamma}_t, 0) \mid \Gamma \right]+E \left[
\ell^q(\hat{\gamma}_{t+1}, 1) \mid \Gamma \right]
\\ &\phantom{{}=}- \left(E \left[ \ell^q(\hat{\gamma}_t,1)
\mid Swap(\Gamma,t) \right]+E \left[ \ell^q(\hat{\gamma}_{t+1}, 0)
\mid Swap(\Gamma,t) \right] \right)
\\
&= q\left( 1- F_{\beta \left(O_{t-1}+1,Z_{t-1}+1\right)}(q) \right) + (1-q) F_{\beta \left(O_{t-1}+1,Z_{t-1}+2\right)}(q) \\
&\phantom{{}=}- \left( (1-q)F_{\beta
\left(O_{t-1}+1,Z_{t-1}+1\right)}(q) + q \left( 1 - F_{\beta
\left(O_{t-1}+2,Z_{t-1}+1\right)}(q) \right) \right) \\
&=
(1-q)F_{\beta \left(O_{t-1}+1,Z_{t-1}+2\right)}(q) + q F_{\beta
\left(O_{t-1}+2,Z_{t-1}+1\right)}(q) - F_{\beta
\left(O_{t-1}+1,Z_{t-1}+1\right)}(q),
\end{align*}
where we used Lemma~\ref{fact:The-expected-reward} for the equality
before last.

By Fact~\ref{fact:RBF_properties}, we have the following recurrence
relations:
\[
F_{\beta(a+1,b)}(x)=F_{\beta(a,b)}(x)-\frac{x^{a}(1-x)^{b}}{a\mathrm{B}(a,b)}
\text{ and }
F_{\beta(a,b+1)}(x)=F_{\beta(a,b)}(x)+\frac{x^{a}(1-x)^{b}}{b\mathrm{B}(a,b)}.
\]
where $B(a,b)$ is the Beta function. Therefore,
%
\begin{align}
\label{eq:tmp swap eq 1}
\begin{split}
R&egret_{TS(q)}^q (\Gamma) - Regret_{TS(q)}^q(Swap(\Gamma,t))\\
&= (1-q)F_{\beta (O_{t-1}+1,Z_{t-1}+2)}(q) + q F_{\beta
(O_{t-1}+2,Z_{t-1}+1)}(q) - F_{\beta (O_{t-1}+1,Z_{t-1}+1)}(q) \\
&= (1-q)\left(F_{\beta(O_{t-1}+1,Z_{t-1}+1)}(q) +
\frac{q^{O_{t-1}+1}(1-q)^{Z_{t-1}+1}}{\left( Z_{t-1}+1 \right)
B\left(O_{t-1}+1,Z_{t-1}+1\right)}\right) \\
& \phantom{{}=}+
 q\left (F_{\beta(O_{t-1}+1,Z_{t-1}+1)} - \frac{q^{O_{t-1}+1}(1-q)^{Z_{t-1}+1}}{( O_{t-1}+1 ) B(O_{t-1}+1,Z_{t-1}+1)}\right) \\
  &\phantom{{}=}-
F_{\beta(O_{t-1}+1,Z_{t-1}+1 )} \\ &=
\frac{q^{O_{t-1}+1}(1-q)^{Z_{t-1}+1}}{B\left(O_{t-1}+1,Z_{t-1}+1\right)}
\left( \frac{1-q}{Z_{t-1}+1} - \frac{q}{O_{t-1}+1} \right),
\end{split}
\end{align}

We now analyse the $sign$ of the terms in Eq. (\ref{eq:tmp swap eq
1}). Since
$\frac{q^{O_{t-1}+1}(1-q)^{Z_{t-1}+1}}{B\left(O_{t-1}+1,Z_{t-1}+1\right)}
> 0$,
\[
sign \Big( Regret_{TS(q)}^q \left(\Gamma\right) - Regret_{TS(q)}^q
\left(Swap\left(\Gamma,t\right)\right) \Big) = sign \left(
\frac{(1-q)}{Z_{t-1}+1} - \frac{q}{O_{t-1}+1} \right).
\]
Thus,
\[
      Regret_{TS(q)}^q \left(\Gamma\right) < Regret_{TS(q)}^q \left(Swap(\Gamma,t)\right) \Longleftrightarrow \\
      \frac{q}{1-q} > \frac{O_{t-1}+1}{Z_{t-1}+1},
\]
  and equality holds iff $\frac{q}{1-q} = \frac{O_{t-1}+1}{Z_{t-1}+1}$.

The second case, where $\gamma_{t}=1$ and $\gamma_{t+1}=0$, is
similar.
\end{proof}
\section{Worst-case regret proofs for $q=\frac{1}{2}$ (Section~\ref{subsec:Worst-case regret half})}

Consider bit sequences $\Gamma = ( \gamma_1, \ldots , \gamma_T )$
with $k $ zeros, where $k\le \frac{T}{2}$ zeros. We first show that
among these bit sequences the ones of largest regret are of the form
$\lbrace 01,10 \rbrace^k1^{T-2k}$. Then, we prove that the regret of
each of these sequences is $\Theta ( \sqrt{k} )$.

\opthalf*

Given the above theorem, to bound the worst case regret of $TS(
\frac{1}{2})$, we can focus on the sequence $W_T^k=\{01\}^k 1^{T-2k}$
and bound $Regret^{1/2}_{TS( \frac{1}{2} )}
(W_T^k)$.

\zoregrethalf*

\begin{proof}
Let $W_T^k = ( w_1, \ldots,w_T )$, where we have: (1) $w_t =0$ for
$t\in A_1=\{2i-1 \mid i\in [k]\}$, (2) $w_t = 1$ for $t\in
A_2=\{2i \mid i\in[k]\}$, and (3) $w_t=1$ for $t\in A_3=\{i \mid i\geq
2k+1\}$.
We  bound the expected number of errors made by $TS ( \frac{1}{2})$
on each of these three subsets.
%
Then, from these bounds we derive a bound on the loss and the regret.

\medskip
\noindent \textbf{The expected number of false positive errors in $A_1$:}
Note that the only errors at times $t\in A_1$ are false positive since $w_t=0$ for these $t$'s.
For  $t \in A_1$  we have that $t=2i-1$, and
$O_{t-1}=Z_{t-1}=i-1$. Hence the algorithm $TS(\frac{1}{2})$ predicts $\hat{\gamma}_{t}= 0$
and  $\hat{\gamma}_{t} = 1$ each with probability of $\frac{1}{2}$ and
\[
\E \left[\mathbb{I} \{\hat{\gamma}_{t} \neq w_{t} \} \mid W_T^k \right] = \frac{1}{2}.
\]
When we sum over $t \in A_1$, we have
\[
\sum\limits _{t=1}^{k} \E \left[\mathbb{I} \{\hat{\gamma}_{t} \neq w_{t} \}  \mid W_T^k  \right] =\frac{k}{2}.
\]

\medskip
\noindent\textbf{The expected number of false negative errors in
$A_2$:}
Note that the only errors at times $t\in A_2$ are false negatives since $w_t=1$.
For  $t \in A_2$  we have $t=2i$, and
$O_{t-1}=i-1$ and $Z_{t-1}=i$. By
Lemma~\ref{fact:The-expected-reward} and Fact~\ref{fact:beta-bin-correlation} we have
\[
\E \left[\mathbb{I} \{\hat{\gamma}_{t} \neq w_{t} \}  \mid W_T^k
\right] = F_{\beta(i,i+1 )}\left( \frac{1}{2} \right)= 1- F_{Bin( 2i,
\frac{1}{2})}(i-1 ).
\]
We can bound $F_{Bin( 2i, \frac{1}{2})}(i-1)$  using Fact~\ref{fact:das}, in the following way
\begin{align*}
F_{Bin( 2i, \frac{1}{2} )}(i-1 ) &= \Pr_{X \sim Bin( 2i, \frac{1}{2}
)}(X \le i ) - \Pr_{X \sim Bin( 2i, \frac{1}{2} )}(X = i ) \\
&= \frac{1}{2} - (1+o(1))\frac{1}{\sqrt{\pi i}}
\end{align*}

Summing over $t \in A_2$ we have,
\[
\sum\limits_{i=1}^{k} \E \left[\mathbb{I} \{\hat{\gamma}_{2i} \neq
w_{2i} \}  \mid W_T^k  \right] = \frac{k}{2} + \sum_{i=1}^{k}
(1+o(1))\frac{1}{\sqrt{\pi i}} = \frac{k}{2} + \Theta ( \sqrt{k} )
\]

\medskip
\noindent \textbf{The expected number of false negative in $A_3$:}
Note that the only errors at times $t\in A_3$ are false negative since $w_t=1$ for these $t$'s.
%
For any $t \in A_3$ we have $Z_t=k$. Therefore,
\[
\E \left[\mathbb{I} \{\hat{\gamma}_{t} \neq w_{t} \}  \mid W_T^k
\right] = F_{\beta(t-k+1,k+1)}\left( \frac{1}{2} \right).
\]
From Theorem~\ref{thm:F_sum_bound-ym} we have
\[
  \sum\limits _{t=2k + 1}^{T} \E \left[\mathbb{I} \{\hat{\gamma}_{t} \neq w_{t} \}  \mid W_T^k  \right] = \sum\limits _{t=2k + 1}^{T} F_{\beta(t-k+1,k+1)}\left( \frac{1}{2} \right) = O (\sqrt{k} ).
  \]

\smallskip

Summing up the errors over $A_1$, $A_2$, and $A_3$ we get that the total number of errors is
\[
\sum\limits _{t=1}^{T} \E \left[\mathbb{I} \{\hat{\gamma}_{t} \neq
w_{t} \}  \mid W_T^k   \right] = \frac{k}{2} +  \left( \frac{k}{2} +
\Theta( \sqrt{k} ) \right) + O (\sqrt{k} ) = k + \Theta( \sqrt{k} )
\]
Recall that the regret is the total loss minus the best static bit prediction.
Since we assume that $k\le \frac{T}{2}$ it is equal to
\begin{equation*}
Regret^{1/2}_{TS( \frac{1}{2} )} (  W_T^k ) = \frac{1}{2}
\sum\limits _{t=1}^{T} \E \left[\mathbb{I} \{\hat{\gamma}_{t} \neq
w_{t} \} \mid W_T^k  \right] - \frac{1}{2} \min \left\lbrace T-k, k
\right\rbrace = \Theta ( \sqrt{k} ).
\end{equation*}
\end{proof}
\section{Best-case regret proofs for $q=\frac{1}{2}$ (Section~\ref{subsec:Best-case regret half})}

We show that for $k \le \frac{T}{2}$, the lowest regret is for the
bit sequence $B_T^k=1^{T-k}0^{k}$. Then, we
prove that its regret is $O(1)$ for any $k\le \frac{T}{2}$ .

\begin{lemma}\label{lem:prefix-swap-half}
For any $\Phi \in \lbrace 0,1 \rbrace^{T-2m}$, $Regret^{1/2}_{TS\left(\frac{1}{2}\right)} (0^m1^m\Phi)=Regret^{1/2}_{TS\left(\frac{1}{2}\right)} (1^m0^m\Phi)$.
\end{lemma}

\begin{proof}
Let $\Gamma^1=( \gamma_1^1,\ldots,\gamma_T^1)=(0^m1^m,\Phi)$ and $\Gamma^2=( \gamma_1^2,\ldots,\gamma_T^2)=(1^m0^m,\Phi)$.
We  show,  using Lemma~\ref{fact:The-expected-reward}, that for each $t\in [T]$, we have $\E [\mathbb{I} \{\hat{\gamma}_{t} = \gamma_{t}^1  \} \mid \Gamma^1 ]=\E[\mathbb{I} \{\hat{\gamma}_{t} = \gamma_{t}^2  \} \mid \Gamma^2 ]$, which implies that $\Gamma^1$ and $\Gamma^2$ have the same expected loss.
Since  static bit prediction also has the same loss on $\Gamma^1$ and $\Gamma^2$ then they have the same regret.

For $t \le m$, by Fact~\ref{fact:beta-symmetry}, we have
\begin{align*}
\E \left[\mathbb{I} \{\hat{\gamma}_{t} = \gamma_{t}^1  \} \mid \Gamma^1 \right] = 1-F_{\beta(1,i+1)}\left(\frac{1}{2}\right)
= F_{\beta(i+1,1)}\left(\frac{1}{2}\right) = \E \left[\mathbb{I} \{\hat{\gamma}_{t}= \gamma_{t}^2  \} \mid \Gamma^2 \right].
\end{align*}
For $m<t \le 2m$ we have,
\begin{align*}
\E \left[\mathbb{I} \{\hat{\gamma}_{t} = \gamma_{t}^1  \} \mid \Gamma^1 \right] = F_{\beta(i+1,m+1)}\left(\frac{1}{2}\right)
= 1-  F_{\beta(m+1,i+1)}\left(\frac{1}{2}\right) = \E \left[\mathbb{I} \{\hat{\gamma}_{t}= \gamma_{t}^2  \} \mid \Gamma^2 \right].
\end{align*}
For $t>2m$ we have $O_t(\Gamma^1)=O_t(\Gamma^2)$ and $Z_t(\Gamma^1)=Z_t(\Gamma^2)$ and thus $\E[\mathbb{I} \{\hat{\gamma}_{t} = \gamma_{t}^1  \} \mid \Gamma^1 ]=\E[\mathbb{I} \{\hat{\gamma}_{t} = \gamma_{t}^2  \} \mid \Gamma^2 ]$.
%
\end{proof}

From that we can induce that $B_T^k$ has the lowest regret on $TS(q)$.

\besthalf*
\begin{proof}
Let $\Gamma=(\gamma_1,\ldots,\gamma_T ) \in \lbrace 0,1 \rbrace^T$
be a bit sequence of length $T$ with $k\le \frac{T}{2}$ zeros such that $\Gamma
\neq 1^{T-k}0^{k}$. We show that there is a bit sequence $\tilde{\Gamma}$, that has the same regret as $\Gamma$, and for some $t\in[T]$ the sequence $Swap(\tilde{\Gamma},t)$ has regret
smaller than $\tilde{\Gamma}$.

Since $\Gamma \neq 1^{T-k}0^{k}$, then either $\Gamma = 0^{k}1^{T-k}$ or it has a prefix of the form $0^m1^n0$ or
$1^n0^m1$, where $n,m > 0$.

First, we look at the case where $\Gamma = 0^{k}1^{T-k}$. By Lemma~\ref{lem:prefix-swap-half}, the sequence $\tilde{\Gamma} = 1^{k}0^{k}1^{T-2k}$ has the same regret as $\Gamma$ and by Lemma~\ref{lem:swap-lemma}, the sequence $Swap (
\tilde{\Gamma}, 2k )$ has regret smaller than the regret of $\tilde{\Gamma}$.

Second, assume $\Gamma$ has a prefix of
$0^m1^n0$ (the case of $1^n0^m1$ is similar).
We have two sub-cases: (a) If $m \ge n$ then $O_{n+m-1}<Z_{n+m-1}$ and
$\gamma_{n+m}=1$, $\gamma_{n+m+1}=0$. By
Lemma~\ref{lem:swap-lemma}, the sequence $Swap ( \Gamma, n+m )$ has
regret lower than $\Gamma$. (b) If $m < n$, by Lemma~\ref{lem:prefix-swap-half}, the bit sequences $\Gamma=(0^m1^m1^{n-m}0,\gamma_{m+n+2},\ldots,\gamma_T )$ and
$\tilde{\Gamma}=(1^m0^m1^{n-m}0,\gamma_{m+n+2},\ldots,\gamma_T )$ have the same regret.
By Lemma~\ref{lem:swap-lemma}, the sequence $Swap (
\tilde{\Gamma}, 2m )$ has regret smaller than the regret of $\tilde{\Gamma}$.

For $k=\frac{T}{2}$, by Lemma~\ref{lem:prefix-swap-half}, both
 $0^{T/2}1^{T/2}$ and $1^{T/2}0^{T/2}$ have the same regret.
\end{proof}

We now bound the regret of $B_T^k=1^{T-k}0^k$.

\zobesthalf*
\begin{proof}
For $t \le T-k$ we have $b_t=1$. Thus
\[
\E \left[\mathbb{I} \{\hat{\gamma}_{t} \neq b_{t} \} \mid B_T^k
\right] = F_{\beta(O_{t-1}+1,Z_{t-1} + 1 )}\left( \frac{1}{2} \right) =
F_{\beta(t,1 )}\left( \frac{1}{2} \right).
\]
Using Fact~\ref{fact:beta-pow}, we have
\[
\E \left[\mathbb{I} \{\hat{\gamma}_{t} \neq b_{t} \} \mid B_T^k
\right] = \left(\frac{1}{2}\right)^t.
\]
This implies that the expected number of false negative errors, in steps $t \le T-k$, is
\[
\sum\limits_{t=1}^{T-k} \E \left[\mathbb{I} \{\hat{\gamma}_{t} \neq
b_{t} \} \mid B_T^k \right] = \sum\limits_{t=1}^{T-k}
\left(\frac{1}{2}\right)^t \le 1.
\]

For $t \ge T-k +1 $ we can have at most $k$ errors so
\[
\sum_{t=T-k+1}^{T} \E \left[\mathbb{I} \{\hat{\gamma}_{t} \neq b_{t}
\} \mid B_T^k \right] \le k.
\]

Therefore, the regret of $TS( \frac{1}{2} )$ on $B_T^k$ is bounded by
\begin{align*}
Regret^{1/2}_{TS( \frac{1}{2} )} ( B_T^k ) &=
\frac{1}{2}\sum\limits_{t=1}^{T} \E \left[\mathbb{I}
\{\hat{\gamma}_{t} \neq b_{t} \} \mid B_T^k \right] - \frac{1}{2}
\min \left\lbrace T-k, k \right\rbrace
\\ &\le \frac{1}{2}(k + 1) - \frac{1}{2}\min \left\lbrace T-k, k \right\rbrace \leq  1.
\end{align*}
\end{proof}

\section{Worst-case regret proofs for a general $q$ (Sections \ref{subsec:wc-seq} and \ref{subsec:wc-reg-q})}

Recall $H^q$,
\begin{equation}
\begin{split}
\forall \Phi \in \left\lbrace 0,1 \right\rbrace^*: H^{q} \left( \Phi \right) = \left\{
  \begin{matrix}
    \{ 0 \} & \frac{O( \Phi ) + 1}{Z( \Phi ) + 1} > \frac{q}{1-q} \\
    \{ 1 \} & \frac{O( \Phi ) + 1}{Z( \Phi ) + 1} < \frac{q}{1-q} \\
    \{ 0,1 \} & \frac{O( \Phi ) + 1}{Z( \Phi ) + 1} = \frac{q}{1-q}
  \end{matrix}
\right. ,
\end{split}
\end{equation}
where $O(\Phi)$ is the total number of $1$s in $\Phi$ and $Z(\Phi)$ is the total number of $0$s in $\Phi$.
For every  sequence $\Gamma = (\gamma_1,\ldots,\gamma_T) \in \{0,1\}^T$ we define  $\nfpre(\Gamma)$ to be the largest index $t$ s.t. $\forall i \in [ t ]: \gamma_i \in H^{q} ( \Gamma_{1:i-1} )$, where $ \Gamma_{1:n}=(\gamma_1,\ldots,\gamma_n)$. We call a bit sequence $\Gamma=(\gamma_1,\ldots,\gamma_T )$ a \textit{\nwcseq} if $\gamma_{\nfpre(\Gamma)+1}=\ldots=\gamma_T$. We define the subsequence $(\gamma_1,\ldots,\gamma_{\nfpre(\Gamma)})$ as the \textit{\perpart} of $\Gamma$ and denote it
$\perpart(\Gamma)$ and the subsequence $(\gamma_{\nfpre(\Gamma)+1},\ldots,\gamma_T)$ as the \textit{\ntailpart}
of $\Gamma$ and denote it
$\ntailpart(\Gamma)$.

For start, we want to bound the number of $0$s and $1$s in the \perpart of a \nwcseq.

\begin{lemma} \label{lem:Z-O-bound}
Fix  a \wcseq $\Gamma=( \gamma_1,\ldots,\gamma_T )$ and let $t \le \nfpre(\Gamma)$. Then, if $\gamma_t=0$ then
$(1-q)t \le Z_t \le (1-q)t + (1-q)$ and $qt - (1-q) \le O_{t} \le qt$, if $\gamma_t=1$ then $(1-q)t - q \le Z_t \le (1-q)t$ and $qt \le O_{t} \le qt + q$.
\end{lemma}

\begin{proof}
The proof is by induction on $t$. For $t=1$ and $q < \frac{1}{2}$ we have that $\frac{q}{1-q} < 1$ and therefore $H^q$ of an empty sequence equals $\{0\}$. Thus, as $t \le \nfpre(\Gamma)$, we must place $\gamma_1 = 0$. In case of such sequence $(1-q) \le 1 \le 2(1-q)$ and $2q - 1 \le 0 \le q$.

By the induction hypothesis for both $\gamma_{t-1}=0$ and $\gamma_{t-1}=1$ we have, $(1-q)(t-1) - q \le Z_{t-1} \le (1-q)(t-1) + (1-q)$ and $q(t-1) - (1-q) \le O_{t-1} \le q(t-1) + q$.

\textbf{Case 1} $\gamma_t=0$. Since $t \le \nfpre(\Gamma)$, we have that $0 \in H^q ( \Gamma_{1:t-1} )$ and therefore $\frac{O_{t-1} + 1}{Z_{t-1} + 1} \ge \frac{q}{1-q}$. Since $O_{t-1}=O_t$ and $Z_{t-1}+1=Z_t$ we get that
\begin{equation} \label{eq:eilon2}
\frac{O_{t} + 1}{Z_{t}} \ge \frac{q}{1-q}\ .
\end{equation}
Since $Z_t+O_t=t$ we can substitute $Z_t=t -O_t$ in Eq. (\ref{eq:eilon2})
and get that $O_{t} \ge qt - (1-q)$. Similarly by substituting $O_t=t -Z_t$ in Eq. (\ref{eq:eilon2}) we get that
 $Z_{t} \le (1-q)t+(1-q)$.
The upper bound on $O_t$ and the lower bound on $Z_t$ follow directly from our assumption:
$Z_{t} = Z_{t-1}+1 \ge (1-q)(t-1) - q  + 1 =(1-q)t$ and $O_{t} = O_{t-1} \le q(t-1) + q=qt$.

\textbf{Case 2} $\gamma_t=1$. Since $t \le \nfpre(\Gamma)$, we have that $1 \in H^q ( \Gamma_{1:t-1} )$ and therefore $\frac{O_{t-1} + 1}{Z_{t-1} + 1} \le \frac{q}{1-q}$. Since $O_{t-1}+1=O_t$ and $Z_{t-1}=Z_t$ we get that
\begin{equation} \label{eq:eilon}
\frac{O_{t}}{Z_{t} + 1} \le \frac{q}{1-q}\ .
\end{equation}
Since $Z_t+O_t=t$ we can substitute $Z_t=t -O_t$ in Eq. (\ref{eq:eilon})
and get that $O_{t} \le qt + q$. Similarly by substituting $O_t=t -Z_t$ in Eq. (\ref{eq:eilon}) we get that
 $Z_{t} \ge (1-q)t - q$.
The lower bound on $O_t$ and the upper bound on $Z_t$ follow directly from our assumption:
 $Z_{t} = Z_{t-1} \le (1-q)(t-1)+(1-q)=(1-q)t$ and $O_{t} = O_{t-1}+1 \ge q(t-1) - (1-q) + 1 = qt$.
\end{proof}

From Lemma~\ref{lem:Z-O-bound} we characterize the \tailpart of a \nwcseq.

\tailfill*
\begin{proof}
Let $j=\nfpre(\Gamma)$.

Consider first the case where
 $Z_T \le (1-q)T - q$ and assume by contradiction that $\ntailpart(\Gamma)$  is not empty and it is filled with zeros.
It follows from this assumption that $Z_T = Z_j+(T-j)$.
By Lemma~\ref{lem:Z-O-bound} we have
that $Z_j \ge (1-q)j-q$, and by combining this inequality with the equality $Z_T = Z_j+(T-j)$
we get that
 $Z_T \ge (1-q)j-q +T-j = T-qj-q$.
On the other hand we assumed that  $Z_T \le (1-q)T - q$. So by combining these upper and lower bounds on $Z_T$ we get that
 $(1-q)T - q \ge T-qj-q$ and thus $j\ge T$. This is a contradiction to the assumption that $\ntailpart(\Gamma)$  is not empty.

Consider now the case where
 $Z_T \ge (1-q)T - q+1$ and assume by contradiction that $\ntailpart(\Gamma)$  is not empty and it is filled with ones.
It follows from this assumption that $O_T = O_j+(T-j)$.
By Lemma~\ref{lem:Z-O-bound} we have
that $O_j \ge qj-(1-q)$, and by combining this inequality with the equality $O_T = O_j+(T-j)$
we get that
 $O_T \ge qj-(1-q) +T-j = T-(1-q)j-(1-q)$.
On the other hand we assumed that  $O_T=T-Z_T \le qT -(1-q)$. So by combining these upper and lower bounds on $O_T$ we get that
 $qT -(1-q) \ge T-(1-q)j-(1-q)$ and thus $j \ge T$. This is a contradiction to the assumption that $\ntailpart(\Gamma)$  is not empty.
\end{proof}

Now we prove that all the worst-case sequences have the largest regret and bound it.

\worstkt*

\begin{proof}
Let $i=\nfpre(\Gamma)+1$.
Since $\Gamma$ is not a \nwcseq,
there is an index $j>i$ such that $\gamma_j\not= \gamma_i$ (since $\ntailpart(\Gamma)$ contains both $0$'s and $1$'s).
Assume $j$ is the smallest index with this property.

\noindent
{\bf Case 1} Assume $\gamma_i=0$ and $\gamma_j=1$. Since $\gamma_i \notin H^q(\Gamma_{1:i-1})$ we have $\frac{O_{i-1} (\Gamma) + 1}{Z_{i-1}(\Gamma) + 1} < \frac{q}{1-q}$. From the definition of $j$ follows that $\gamma_{i}= \gamma_{i+1} = \ldots =\gamma_{j-1} = 0$ and thus $\frac{O_{j-2}(\Gamma) + 1}{Z_{j-2}(\Gamma) + 1} \le \frac{O_{i-1}(\Gamma) + 1}{Z_{i-1}(\Gamma) + 1} < \frac{q}{1-q}$. By Lemma \ref{lem:swap-lemma}, the sequence $Swap ( \Gamma, j-1 )$ has a regret larger than $\Gamma$.

\noindent
{\bf Case 2} Assume $\gamma_i=1$ and $\gamma_j=0$. Since $\gamma_i \notin H^q(\Gamma_{1:i-1})$ we have $\frac{O_{i-1} (\Gamma) + 1}{Z_{i-1}(\Gamma) + 1} > \frac{q}{1-q}$. From the definition of $j$ follows that $\gamma_{i}= \gamma_{i+1} = \ldots =\gamma_{j-1} = 1$ and thus $\frac{O_{j-2}(\Gamma) + 1}{Z_{j-2}(\Gamma)  + 1} \ge \frac{O_{i-1}(\Gamma)  + 1}{Z_{i-1}(\Gamma )+ 1} > \frac{q}{1-q}$. By Lemma \ref{lem:swap-lemma}, the sequence $Swap ( \Gamma, j-1 )$ has a regret larger than $\Gamma$.
\end{proof}

Theorem \ref{thm:worst-given-K-T} implies that any
sequence of largest regret is a \nwcseq.
Next we prove that all \wcseqs of length $T$ with $k$ zeros  have the same regret.

\sameregretq*
\begin{proof}
Assume by contradiction that there are two
\wcseqs such that \\$Regret^q_{TS(q)} (\Gamma^1)$ $=r_1$, $Regret^q_{TS(q)} (\Gamma^2)=r_2$ and $r_1\not= r_2$.
We assume further that $\Gamma^1$ and $\Gamma^2$
have the longest common prefix among all \wcseqs of length $T$ with $k$ zeros and regret $r_1$ and $r_2$, respectively.

Since $\Gamma^1$ and $\Gamma^2$ both have $k$ zeros then by
Corollary \ref{cor:tail-fill} their \ntailpart{}s are filled with the same bit. It follows that $\nheadpart(\Gamma^1)\not= \nheadpart(\Gamma^2)$.
Assume without loss of generality that $\nheadpart(\Gamma^2)$ is not shorter than $\nheadpart(\Gamma^1)$.
We claim that $\nheadpart(\Gamma^1)$ is not a prefix of $\Gamma^2$. This follows since otherwise $\Gamma^1$ and
$\Gamma^2$ cannot both have $k$ zeros.

It follows that there exists an index $t\le \nfpre(\Gamma^1)$ such that $\gamma^1_t \not= \gamma^2_t$.
Let $t$  be the smallest such index. Since $\Gamma^1_{1:t-1}=\Gamma^2_{1:t-1}$ we have that
 $\frac{O_{t-1}( \Gamma^1 ) + 1}{Z_{t-1}( \Gamma^1 ) + 1}=\frac{O_{t-1}( \Gamma^2 ) + 1}{Z_{t-1}( \Gamma^2 ) + 1} = \frac{q}{1-q}$.
 Assume  that $\gamma^{1}_t=0$ and $\gamma^{2}_t=1$. Therefore, there is an index $t'>t$ such that
$\gamma^{1}_{t'}=1$ and $\gamma^{2}_{t'}=0$.
Since the tails of both sequences are filled with the same bit then this implies that
$ t'\le \nfpre(\Gamma^2)$ and therefore since $t+1\le t'$ we have that
$ t+1\le \nfpre(\Gamma^2)$.

Since $\gamma^{2}_t=1$ we have that  $\frac{O_{t}(\Gamma^2) + 1}{Z_{t}(\Gamma^2) + 1} > \frac{O_{t-1}(\Gamma^2) + 1}{Z_{t-1}(\Gamma^2) + 1} = \frac{q}{1-q}$, and
since $ t+1\le \nfpre(\Gamma^2)$ we must have that
 $\gamma^{2}_{t+1} = 0$. By Lemma \ref{lem:swap-lemma}, $Regret^q_{TS(q)} (\Gamma^2 ) = Regret^q_{TS(q)} \left(Swap(\Gamma^2, t ) \right) = r_2$.
It is easy to check that
 $Swap(\Gamma^2, t )$ is still a \wcseq and
 since it has a longer common prefix with $\Gamma^1$ we get a contradiction to the choice of $\Gamma^1$ and $\Gamma^2$.

The case where  $\gamma^{1}_t=1$ and $\gamma^{2}_t=0$ is analogous.
\end{proof}

Let $W_T^k=(w_1,\ldots,w_T) \in \{ 0,1 \}^T$
be a \wcseq with $k$ zeros such that for all $t\le \nfpre(W_T^k)$
with $\frac{O_{t-1}+1}{Z_{t-1}+1} = \frac{q}{1-q}$ we have
$\gamma_t = 0$.
 Since
 by Lemma \ref{lem:same-regretq} all the \wcseqs with the same number of zeros have the same regret, we can focus on
bounding the regret of $W_T^k$.

\zoregretq*
\begin{proof}
We first consider the case that $k \le (1-q)T - q$.  We partition
$W_T^k$ into the following sets (1) $ A_1=\{t \mid t\in
[\nfpre(W_T^k)] \text{ and } w_t = 0\}$, (2) $A_2=\{t \mid t\in
[\nfpre(W_T^k)] \text{ and } w_t = 1\}$, and (3) $A_3=\{t \mid t\geq
\nfpre(W_T^k)+1\}$.
We  bound the expected number of errors made by $TS(q)$ on each of
these three subsets. Then, from these bounds we derive a bound on
the loss and the regret.

\medskip
\noindent \textbf{The expected number of false positive errors in $A_1$:}
Note that the only errors at times $t\in A_1$ are false positive since $w_t=0$ for these $t$'s. Therefore, by Lemma~\ref{fact:The-expected-reward} and Fact~\ref{fact:beta-bin-correlation} we have
\begin{align} \label{eq:binq}
  \E \left[\mathbb{I} \{\hat{\gamma}_{t} \neq w_{t} \} \mid W_T^k \right] &= 1 - F_{\beta(O_{t-1}+1,Z_{t-1} + 1)}(q) = F_{\beta(Z_{t-1}+1,O_{t-1} + 1)}(1-q) \nonumber \\&=
  1-F_{Bin( t, 1-q )}(Z_{t-1} ) = 1-F_{Bin( t, 1-q )}(Z_{t}-1).
\end{align}
By the definition of $A_1$, $t\le \nfpre(W_T^k)$, and therefore by Lemma~\ref{lem:Z-O-bound}, $(1-q)t \le Z_t$. Thus, $t \le \frac{Z_t}{1-q} \le \frac{Z_t+1-1+q}{1-q} = \frac{Z_t+1}{1-q} - 1 \le \floor*{\frac{Z_t+1}{1-q}}$. Let $m=\floor*{\frac{Z_{t}+1}{1-q}}$ and $X \sim Bin\left( m, 1-q \right)$.
We can bound the right side of Eq. (\ref{eq:binq}) as follows.
\begin{align} \label{eq:bunq}
F_{Bin( t, 1-q )} (Z_{t}-1) &\ge F_{Bin\left( m, 1-q \right)} (Z_{t}-1)
\nonumber \\ &= \Pr (X \le Z_t + 1 )- \Pr (X = Z_t + 1) - \Pr (X = Z_t).
\end{align}

We now bound the different probabilities in Eq. (\ref{eq:bunq}).
Since $X$ is a Binomial random variable, its median is $\floor*{m(1-q)}=Z_t$ or $\ceil*{m(1-q)}=Z_t+1$ and thereby
\begin{equation} \label{eq:A1_Z10}
\Pr (X \le Z_t +1) \ge \frac{1}{2}.
\end{equation}

For any $Z_t \ge \frac{2(1-q)}{q} - 1$, we bound $\Pr (X = Z_{t}+1)$ by Lemma~\ref{lem:binom-bound} as follows
\begin{equation} \label{eq:A1_Z1}
\Pr (X = Z_{t}+1) = O\left(\frac{1}{\sqrt{q Z_t}}\right).
\end{equation}
The probability $\Pr (X = Z_t)$ is bounded using the previous equality,
\begin{align} \label{eq:A1_Z2}
\frac{\Pr (X = Z_t)}{\Pr (X = Z_{t}+1)} &= \frac{q(Z_t+1)}{(1-q)(m-Z_t)} \le \frac{q(Z_t+1)}{(1-q)(\frac{Z_t+1}{1-q} - 1-Z_t)} \nonumber \\&= \frac{q(Z_t+1)}{(1-q)(\frac{Z_t+1 - (1-q)(Z_t+1)}{1-q})} = \frac{q(Z_t+1)}{q(Z_t+1)} = 1.
\end{align}
Therefore by using Eq. (\ref{eq:A1_Z1}) and (\ref{eq:A1_Z2}) we have
\begin{equation} \label{eq:A1_Z4}
\Pr (X = Z_t) = O\left(\frac{1}{\sqrt{q Z_t}}\right).
\end{equation}
By substituting Eq. (\ref{eq:bunq}-\ref{eq:A1_Z1},\ref{eq:A1_Z4}) into (\ref{eq:binq}) we get that for $Z_t \ge \frac{2(1-q)}{q} - 1$
\begin{equation} \label{eq:miss-bound1}
\E \left[\mathbb{I} \{\hat{\gamma}_{t} \neq w_{t} \} \mid W_T^k \right] \le
\frac{1}{2} + O\left(\frac{1}{\sqrt{q Z_t}}\right).
\end{equation}
For $Z_t < \frac{2(1-q)}{q} - 1$ we assume the worst-case to have
\begin{equation} \label{eq:miss-bound2}
\E \left[\mathbb{I} \{\hat{\gamma}_{t} \neq w_{t} \} \mid W_T^k \right] \le 1.
\end{equation}

Notice that since $k \le (1-q)T - q$, by Corollary~\ref{cor:tail-fill} there are no zeros in the \ntailpart .Thus, all the zeros of $W_T^k$ are in $A_1$. Thus, we use Eq. (\ref{eq:miss-bound1}-\ref{eq:miss-bound2}) to sum over all $t\in A_1$.
\begin{align} \label{eq:A1-total}
\sum\limits_{t\in A_1} &\E \left[\mathbb{I} \{\hat{\gamma}_{t} \neq w_{t} \} \mid W_T^k \right]
\nonumber \\ &= \sum\limits_{\{t\in A_1 \mid  Z_t < \frac{2(1-q)}{q}-1\}} \E \left[\mathbb{I} \{\hat{\gamma}_{t} \neq w_{t} \} \mid W_T^k \right] + \sum\limits_{\{t\in A_1 \mid Z_t \ge \frac{2(1-q)}{q}-1\}} \E \left[\mathbb{I} \{\hat{\gamma}_{t} \neq w_{t} \} \mid W_T^k \right]  \nonumber
\\&\leq \frac{2(1-q)}{q}-1 + \sum\limits_{t\in A_1} \left( \frac{1}{2} + O\left( \frac{1}{\sqrt{2\pi q Z_t}} \right)\right) \
 \nonumber \\ &\le O \left( \frac{1-q}{q}\right) + \frac{k}{2} + \sum\limits_{i=1}^{k} O\left( \frac{1}{\sqrt{2\pi q i}} \right)
= \frac{k}{2} + O \left( \sqrt{ \frac{k}{q}} +  \frac{1-q}{q} \right) .
\end{align}

\medskip
\noindent\textbf{The expected number of false negative errors in
$A_2$:}
Note that the only errors at times $t\in A_2$ are false negative since $w_t=1$. Therefore, by Lemma~\ref{fact:The-expected-reward} and Fact~\ref{fact:beta-bin-correlation} we have
\begin{equation} \label{eq:A2-1}
\E \left[I \{\hat{\gamma}_t \neq w_{t} \} \mid W_T^k \right] = F_{\beta (O_{t-1}+1,Z_{t-1} + 1)}(q) = 1- F_{Bin( t, q )}(O_{t-1})=1- F_{Bin( t, q )}(O_{t}-1).
\end{equation}
By the definition of $A_2$, $t\le \nfpre(W_T^k)$, and therefore by Lemma~\ref{lem:Z-O-bound}, $qt \le O_{t}$. Thus, $t \le \frac{O_t}{q} \le \frac{O_t+1-q}{q} = \frac{O_t+1}{q} - 1 \le \floor*{\frac{O_t+1}{q}}$. Let $m=\floor*{\frac{O_{t}+1}{q}}$ and $X \sim Bin\left( m, q \right)$. We can continue and bound the right side of Equation (\ref{eq:A2-1}) as follows.
\begin{align} \label{eq:A2-2}
F_{Bin( t, q )} (O_{t}-1) &\geq F_{Bin\left( m, q \right)} (O_{t}-1)
\nonumber \\ &= \Pr (X \le O_{t} + 1 ) - \Pr (X = O_{t} + 1 ) - \Pr (X = O_{t}).
\end{align}
Note that we have analogous bounds to the previous case of $A_1$, since by substituting $Z_t$ and $1-q$ by $O_t$ and $q$ respectively in Eq. (\ref{eq:binq},\ref{eq:bunq}) we get Eq. (\ref{eq:A2-1},\ref{eq:A2-2}). Thereby,
\begin{equation} \label{eq:A2-3}
\E \left[I \{\hat{\gamma}_t \neq w_{t} \} \mid W_T^k \right] \le \left\{
\begin{matrix}
\frac{1}{2} + O \left(\frac{1}{\sqrt{(1-q) O_t}}\right) & O_t \ge \frac{2q}{1-q}-1 \\
1 & O_t < \frac{2q}{1-q}-1
\end{matrix}
\right. .
\end{equation}

Since $\nperpart(W^k_T)$ contains all the zeros in $W^k_T$ we have $Z_{\nfpre(W^k_T)}=k$. By using Lemma~\ref{lem:Z-O-bound} we get that $(1-q)\nfpre(W^k_T) -q \le Z_{\nfpre(W^k_T)}$ and thus
$\nfpre(W^k_T) \le \frac{k+q}{1-q}$.  Therefore, $O_{\nfpre(W^k_T)} = \nfpre(W^k_T)-Z_{\nfpre(W^k_T)} \le \frac{k+q}{1-q} - k \le \frac{q}{1-q}k+1$.

Let $n=\ceil*{\frac{q}{1-q}k}+1$. By Eq. (\ref{eq:A2-3}), we sum over all $t\in A_2$ to have
\begin{align} \label{eq:A2-total}
\sum\limits_{t \in A_2} &\E \left[\mathbb{I} \{\hat{\gamma}_{t} \neq
w_{t} \}  \mid W_T^k  \right] \le \frac{2q}{1-q}-1 + \sum_{\{t\in A_2 \mid O_t \ge \frac{2q}{1-q}-1\}}
O \left(\frac{1}{2}+ \frac{1}{\sqrt{2\pi(1-q) O_t}}\right) \\&\le \frac{2q}{1-q}-1 + \frac{n}{2} + \sum_{i=1}^{n}
O \left(\frac{1}{\sqrt{(1-q) i}} \right)
= \frac{n}{2} + O \left( \sqrt{\frac{n}{1-q}} + \frac{q}{1-q} \right)
\nonumber \\& \le \frac{\frac{q}{1-q}k+2}{2} + O \left( \sqrt{\frac{\frac{q}{1-q}k+2}{1-q}} + \frac{q}{1-q} \right)
= \frac{qk}{2(1-q)} + O \left( \frac{\sqrt{qk}}{1-q} + \frac{q}{1-q} \right), \nonumber
\end{align}
where the one before last inequality follows from substitution of $n=\ceil*{\frac{q}{1-q}k}+1 \le \frac{q}{1-q}k+2$.

\medskip
\noindent \textbf{The expected number of false negative in $A_3$:}
By Corollary~\ref{cor:tail-fill} the only errors at times $t\in A_3$ are false negative since $w_t=1$. For any $t \in A_3$ we have $Z_t=k$. Therefore,
\[
\E \left[\mathbb{I} \{\hat{\gamma}_{t} \neq w_{t} \}  \mid W_T^k
\right] = F_{\beta(t-k+1,k+1)}( q ).
\]
From Lemma~\ref{lem:Z-O-bound}, $(1-q)\nfpre(W^k_T) +(1-q) \ge Z_{\nfpre(W^k_T)} = k$ and thus $\nfpre(W^k_T) \ge \frac{k}{1-q}-1$. From Theorem~\ref{thm:F_sum_bound-generalized} we have
\begin{align} \label{eq:A3-total}
  \sum\limits _{t=\floor*{\frac{k}{1-q}}-1}^{T} \E \left[\mathbb{I} \{\hat{\gamma}_{t} \neq w_{t} \}  \mid W_T^k  \right]
  &= \sum\limits_{t=\floor*{\frac{k}{1-q}}-1}^{T} F_{\beta(t-k+1,k+1)}( q )
  \le \sum\limits_{i=\floor*{\frac{qk}{1-q}} - 2}^{\infty} F_{\beta(i+1,k+1)}( q )
  \nonumber \\&\le 3 + \sum\limits_{i=\floor*{\frac{qk}{1-q}} + 1}^{\infty} F_{\beta(i+1,k+1)}( q )
  = O\left(\sqrt{qk}\right),
\end{align}
where the inequality follows from $t-k = \floor*{\frac{k}{1-q}}-1 -
k \ge  \frac{k}{1-q}- k - 2 =\frac{qk}{1-q} - 2=i$.

Since $k \le (1-q)T - q$, the best static bit predictor is
\[
static^q(W^k_T) = \min\{ (1-q)(T-k), qk\} = qk.
\]
By using Eq. (\ref{eq:A1-total}), (\ref{eq:A2-total}) and (\ref{eq:A3-total}), the regret is the total loss minus the best static bit prediction
\begin{align*}
Regret^q_{TS(q)} (W_T^k )
&= \sum\limits_{t=1}^{T} \E_{\hat{\gamma}_t \sim TS(q)} \left[
\ell^q(\hat{\gamma}_t, w_t) \mid W^k_T \right] - static^q
\left( W^k_T \right)
\\&=q\left(\frac{k}{2} + O \left( \sqrt{\frac{k}{q}}+  \frac{1-q}{q}  \right) \right)+(1-q)\left(\frac{qk}{2(1-q)} + O \left( \frac{\sqrt{qk}}{1-q} +\frac{q}{1-q} \right)\right) \\
&\phantom{{}==}+ (1-q)O(\sqrt{qk}) - \min \left\lbrace (1-q)(T-k), qk \right\rbrace \\
&= O \left( \sqrt{qk}\right).
\end{align*}

We now look at the regret for $k \ge (1-q)T-q$. In this proof, we
split the calculations into $A_1,A_2$ and $A_3$ as in
the prior part.

\medskip
\noindent \textbf{The expected number of false positive errors in
$A_1$:} At each $t\in A_1$ the expected errors are bounded in the
same way as in the previous case. The only change is the size of
$A_1$. Notice that since $k > (1-q)T - q$, by
Corollary~\ref{cor:tail-fill} all ones of $W_T^k$ are in $A_1$. By
the definition of $A_1$, $t \le \nfpre(W_T^k)$, and therefore by
Lemma~\ref{lem:Z-O-bound}, $qt-(1-q) \le O_{\nfpre(W_T^k)}=T-k$ and
thus $t \le \frac{T-k+1}{q}$ . From Lemma~\ref{lem:Z-O-bound} we
also conclude that$Z_{\nfpre(W_T^k)} \le (1-q)\nfpre(W_T^k) + (1-q)
\le (1-q)\frac{T-k+1}{q} + (1-q)$. In total, $|A_1| \le
(1-q)\frac{T-k+1}{q} + (1-q)$. Thereby, the expected number of
errors in $A_1$ is bounded by $\frac{\frac{1-q}{q}(T-k)}{2} + O
\left( \frac{\sqrt{(1-q)(T-k)}}{q} + \frac{1-q}{q} \right)$.

\medskip
\noindent\textbf{The expected number of false negative errors in
$A_2$:} At each $t\in A_2$ the expected errors are bounded in the
same way as in the previous case. The only change is the size of
$A_2$, which equals to $T-k$ since from
Corollary~\ref{cor:tail-fill} all the ones of $W^k_T$ are in
$\nperpart(W^k_T)$. Thus we have that the expected number of errors
is bounded by $ \frac{T-k}{2} + O \left( \sqrt{\frac{T-k}{1-q}} +
\frac{q}{1-q} \right)$.

\medskip
\noindent \textbf{The expected number of false negative in $A_3:$}
By Corollary~\ref{cor:tail-fill} the only errors at times $t\in A_3$
are false negative since $w_t=0$. For any $t \in A_3$ we have
$O_t=T-k$. Therefore,
\[
  \E \left[\mathbb{I} \{\hat{\gamma}_{t} \neq w_{t} \} \mid W_T^k \right] = 1-F_{\beta(T-k+1,t-(T-k)+1)}(q)= F_{\beta(t-(T-k)+1,T-k+1)}(1-q).
\]
From Lemma~\ref{lem:Z-O-bound}, $q\nfpre(W^k_T) + q \ge O_{\nfpre(W^k_T)} = T-k$ and thus
$\nfpre(W^k_T) \ge \frac{T-k}{q}-1$. From Theorem~\ref{thm:F_sum_bound-generalized}, since $1-q\ge \frac{1}{2}$, we have
\begin{align*}
  \sum\limits _{t=\floor*{\frac{T-k}{q}}-1}^T \E \left[\mathbb{I} \{\hat{\gamma}_{t} \neq w_{t} \} \mid W_T^k \right] &= \sum\limits _{t=\floor*{\frac{T-k}{q}}-1}^{T} F_{\beta(t-(T-k)+1,T-k+1)}(1-q) \\ &\le 3+\sum\limits _{i = \floor*{\frac{1-q}{q}(T-k)}+1}^{\infty} F_{\beta(i+1,T-k+1)}(1-q) \\ &=
O\left( \frac{\sqrt{(1-q)(T-k+1)}}{q}+\frac{1-q}{q}e^{-\frac{1}{4(1-q)}(T-k+1)}+\frac{1-q}{q} \right).
\end{align*}

Since $k > (1-q)T - q$, the best static bit predictor is
\[
static^q(W^k_T) = \min\{ (1-q)(T-k), qk\} = (1-q)(T-k).
\]
Hence, the regret in the case is
\begin{align*}
Regret^q_{TS(q)} \left(W_T^k \right)
&= \sum\limits_{t=1}^{T} \E_{\hat{\gamma}_t \sim TS(q)} \left[
\ell^q(\hat{\gamma}_t, w_t) \mid W^k_T \right] - static^q
\left( W^k_T \right)
\\&=q\left(\frac{\frac{1-q}{q}(T-k)}{2} + O \left( \frac{\sqrt{(1-q)(T-k)}}{q} + \frac{1-q}{q} \right)\right)
\\ &\phantom{{}==}+ (1-q) \left( \frac{T-k}{2} + O \left( \sqrt{\frac{T-k}{1-q}} + \frac{q}{1-q} \right) \right)
\\ &\phantom{{}==}+qO\left( \frac{\sqrt{(1-q)(T-k+1)}}{q}+ \frac{1-q}{q}e^{-\frac{1}{4(1-q)}((T-k+1)+1)}+\frac{1-q}{q} \right)
\\ &\phantom{{}==}-\min \left\lbrace (1-q)(T-k), qk \right\rbrace \\
&= O \left( \sqrt{(1-q)(T-k)} \right)
\end{align*}
\end{proof}
\finalregretboundq*
\begin{proof}
Fix $q \in \left[\frac{1}{2},1\right]$ and a bit sequence $\Gamma =( \gamma_1,\ldots,\gamma_T )$. We show that $Regret^q_{TS(q)} \left( \Gamma \right)=Regret^q_{TS(1-q)} \left( \bar{\Gamma} \right)$. At each step $t\in \left[ T \right]$, $O_t \left(\Gamma\right)=Z_t \left(\bar{\Gamma}\right)$. Therefore by Fact \ref{fact:beta-symmetry} we have
\begin{align*}
\E_{\hat{\gamma}_t \sim TS(q)} \left[\mathbb{I} \{\hat{\gamma}_{i_t} = 1 \} \mid \Gamma \right] &= \Pr_{x_t \sim \beta ( O_{t-1}(\Gamma) + 1, Z_{t-1}(\Gamma) + 1 )} ( x_t > q )
\\ &= \Pr_{x_t \sim \beta ( Z_{t-1}(\Gamma) + 1, O_{t-1}(\Gamma) + 1 )} ( x_t < 1-q)
\\ &= \Pr_{x_t \sim \beta \left( O_{t-1}\left(\bar{\Gamma} \right) + 1, Z_{t-1}\left(\bar{\Gamma} \right) + 1 \right)} (x_t < 1-q)
\\ &= \E_{\hat{\gamma}_t \sim TS(1-q)} \left[\mathbb{I} \{\hat{\gamma}_{i_t} = 0 \} \mid \bar{\Gamma} \right].
\end{align*}
The benchmarks are the same as,
\begin{align*}
static_q \left( \Gamma \right) &= \min \lbrace \left( 1-q \right) O_T \left( \Gamma \right), q Z_T \left( \Gamma \right) \rbrace
\\ &= \min \left\lbrace q O_T \left( \bar{\Gamma} \right), \left( 1-q \right) Z_T \left( \bar{\Gamma} \right) \right\rbrace = static_{1-q} \left( \bar{\Gamma} \right).
\end{align*}
We conclude that $Regret^q_{TS(q)} \left( \Gamma \right)=Regret^{1-q}_{TS(1-q)} \left( \bar{\Gamma} \right)$.
%
\end{proof}
\worstqlarge*
\begin{proof}
Assume $q\in \left[0, \frac{1}{2} \right]$. From Theorem
\ref{thm:worst-given-K-T} the bit sequences that generate the
largest regret, with $k$ zeros, are \nwcseqs. Theorem
\ref{thm:0-1-regret} shows that the regret of these bit sequences is
\[
\left\{
                \begin{matrix}
                  O \left( \sqrt{qk} \right) & k \le (1-q)T - q \\
                  O \left( \sqrt{(1-q)(T-k)} \right) & \textnormal{otherwise}
                \end{matrix}
              \right. .
\]
Thus, the worst-case regret over all $k$'s is
\[
\max \left\lbrace O \left( \sqrt{q(1-q)T}\right), O \left( \sqrt{(1-q)(T-(1-q)T)}\right) \right\rbrace=O \left( \sqrt{q(1-q)T}\right).
\]

For $q\in \left[\frac{1}{2}, 1 \right]$,
Lemma~\ref{cor:final-regret-bound} with Theorem \ref{thm:0-1-regret}
gives us the same regret of $O \left(\sqrt{q(1-q)T}\right)$.
\end{proof}

\section{Best-case regret proofs for a general $q$ (Section \ref{subsec:best-reg-q})}
\zobestregretq*

\begin{proof}
First we calculate the loss of $\Gamma_1=1^n0^m$.
For $t \le n$ we have $\gamma_t=1$. Thus, by using Lemma~\ref{fact:The-expected-reward}, 
\[
\E \left[\mathbb{I} \{\hat{\gamma}_{t} \neq \gamma^{(1)}_{t} \} \mid \Gamma_1
\right] 
= F_{\beta(O_{t-1}+1,Z_{t-1} + 1 )}(q) =
F_{\beta(t,1 )}(q).
\]
Using Fact~\ref{fact:beta-pow}, we have
\[
F_{\beta(t,1 )}(q) = q^t.
\]
This implies that the expected number of false negative errors, in steps $t \le n$, is
\[
\sum\limits_{t=1}^{n} \E \left[\mathbb{I} \{\hat{\gamma}_{t} \neq
\gamma^{(1)}_{t} \} \mid \Gamma_1 \right] = \sum\limits_{t=1}^{n}
q^t \le \frac{1}{1-q}.
\]
For $t \ge n +1 $ we can have at most $m$ errors so
\[
\sum_{t=n+1}^{T} \E \left[\mathbb{I} \{\hat{\gamma}_{t} \neq \gamma^{(1)}_{t}
\} \mid \Gamma_1 \right] \le m.
\]
Therefore, the expected loss of $TS( q )$ on $\Gamma_1$ is bounded by
\begin{align} \label{eq:0-1-best-1}
\sum\limits_{t=1}^{T} E \left[
\ell^q(\hat{\gamma}_t, \gamma^{(1)}_t) \mid \Gamma_1 \right]
&= (1-q)\sum\limits_{t=1}^{n} \E \left[\mathbb{I}
\{\hat{\gamma}_{t} \neq \gamma^{(1)}_{t} \} \mid \Gamma_1 \right] + q\sum\limits_{t=n+1}^{n+m} \E \left[\mathbb{I}
\{\hat{\gamma}_{t} \neq \gamma^{(1)}_{t} \} \mid \Gamma_1 \right] 
\nonumber \\ &\le (1-q)\frac{1}{1-q}+qm=1+qm.
\end{align}

Analogously, we bound the expected loss of $TS( q )$ on $\Gamma_2=0^m1^n$ by
\begin{align} \label{eq:0-1-best-2}
\sum\limits_{t=1}^{T} E \left[
\ell^q(\hat{\gamma}_t, \gamma^{(2)}_t) \mid \Gamma_2 \right]
\le 1+(1-q)n.
\end{align}

The benchmark of the two sequences is the same and equals
\begin{align*}
static^q(\Gamma_1) = static^q(\Gamma_2) = \min \{ qm, (1-q)n \}.
\end{align*}
Therefore, if $\min \{ qm, (1-q)n \} = qm$ then by Eq. (\ref{eq:0-1-best-1})
\[
Regret^q_{TS(q)} (\Gamma_1) \le 1+qm -qm = 1.
\]
Otherwise $\min \{ qm, (1-q)n \} = (1-q)n$ and by Eq. (\ref{eq:0-1-best-2})
\[
Regret^q_{TS(q)} (\Gamma_2) \le 1+(1-q)n -(1-q)n = 1.
\]
\end{proof}

\section{Binomial
coefficient approximations}

We use the following well known approximation of the Binomial
coefficient using Stirling's approximation. (see for example, \cite{binomial})

\begin{fact}
\label{fact:das}
For every $m \in \mathbb{N}^+$ and $n \le m$ we have
\[
\binom{m}{n} = (1+o(1))\sqrt{\frac{m}{2\pi n (m-n)}} \left(\frac{m}{n}\right)^n \left(\frac{m}{m-n}\right)^{m-n} .
\]
\end{fact}

From the fact above we conclude the following lemma.

\begin{lemma}
\label{lem:binom-bound}
For every constant $p\in(0,1)$ and $n \ge \frac{2p}{1-p}$ ,  we have
\begin{align*}
\Pr_{X\sim Bin(\floor*{\frac{n}{p}},p)} ( X=n )= O \left(\frac{1}{\sqrt{(1-p) n}} \right).
\end{align*}
\end{lemma}

\begin{proof}
Let $m=\floor*{\frac{n}{p}}$. We bound $\binom{m}{n}$ using Fact~\ref{fact:das} as follows
\begin{align*}
\binom{m}{n} &= (1+o(1))\sqrt{\frac{m}{2\pi n (m-n)}} \left(\frac{m}{n}\right)^n \left(\frac{m}{m-n}\right)^{m-n}.
\end{align*}
From the definition of floor $\exists \omega \in [0,1): m = \frac{n}{p} - \omega$ and therefore
\begin{align*}
\binom{m}{n} &= (1+o(1))\sqrt{\frac{\frac{n}{p} -\omega}{2\pi n (\frac{n}{p} -\omega-n)}} \left(\frac{\frac{n}{p} -\omega}{n}\right)^n \left(\frac{\frac{n}{p} -\omega}{\frac{n}{p} -\omega-n}\right)^{\frac{n}{p} - \omega-n}
\nonumber \\&= O(1) \sqrt{\frac{\frac{n-p \omega}{p}}{n (\frac{(1-p)n-p\omega}{p} )}} \left(\frac{\frac{n-p \omega}{p}}{n}\right)^n \left(\frac{\frac{n-p\omega}{p}}{\frac{(1-p)n-p \omega}{p}}\right)^{\frac{n}{p} - \omega-n}
\nonumber \\&= O(1) \sqrt{\frac{n-p \omega}{n ((1-p)n-p\omega)}} \left(\frac{n-p \omega}{pn}\right)^n \left(\frac{n-p\omega}{(1-p)n-p \omega}\right)^{\frac{n}{p} - \omega-n}.
\end{align*}
Since $0\le p\omega < p$ we have
\begin{align} \label{eq:binom-tmp-1}
\binom{m}{n} &\le O(1) \sqrt{\frac{n}{n ((1-p)n-p)}} \left(\frac{n}{pn}\right)^n \left(\frac{n}{(1-p)n-p}\right)^{\frac{n}{p} - \omega-n}
\nonumber \\&= O(1) \sqrt{\frac{1}{(1-p)n-p}} \left(\frac{1}{p}\right)^n \left(\frac{n}{(1-p)n-p}\right)^{\frac{n}{p} - \omega-n}.
\end{align}
Since $n \ge \frac{2p}{1-p}$ we get that $\frac{(1-p)n}{2} \ge p$ and therefore $(1-p)n-p \ge \frac{(1-p)n}{2}$. Thus, by using Eq. (\ref{eq:binom-tmp-1}),
\begin{equation} \label{eq:binom-tmp-2}
\binom{m}{n} \le  O(1) \sqrt{\frac{2}{(1-p)n}} \left(\frac{1}{p}\right)^n \left(\frac{n}{(1-p)n-p}\right)^{\frac{n}{p} - \omega-n}.
\end{equation}
We bound $\left(\frac{n}{(1-p)n-p}\right)^{\frac{n}{p} - \omega-n}$ as follow
\begin{align} \label{eq:binom-tmp-3}
\left(\frac{n}{(1-p)n-p}\right)^{\frac{n}{p} - \omega-n} &= (1-p)^{-(\frac{n}{p} - \omega-n)} \left(\frac{(1-p)n}{(1-p)n-p}\right)^{\frac{n}{p} - \omega-n} 
\nonumber \\&= (1-p)^{-(\frac{n}{p} - \omega-n)}\left(\frac{1}{1-\frac{p}{(1-p)n}}\right)^{\frac{n}{p} - \omega-n}
\nonumber \\&\le (1-p)^{-(\frac{n}{p} - \omega-n)}\frac{1}{\left(1-\frac{p}{(1-p)n}\right)^{\frac{(1-p)n}{p}}}
\le 4(1-p)^{-(m-n)}.
\end{align}
where the last inequality holds as $\left(1-\frac{p}{(1-p)n}\right)^{\frac{(1-p)n}{p}}$ is a monotonic increasing function and since $n \ge \frac{2p}{1-p}$, the function has a minimum at $n=\frac{2p}{1-p}$.

From Eq. (\ref{eq:binom-tmp-2},\ref{eq:binom-tmp-3}) we have
\[
\Pr_{X\sim Bin(m,p)} ( X=n ) = \binom{m}{n} p^n (1-p)^{m-n}
= O \left(\frac{1}{\sqrt{(1-p) n}} \right).
\]
\end{proof}

\end{document}